\pdfoutput=1

\documentclass[11pt]{article}

\usepackage[final]{acl}

\usepackage{times}
\usepackage{latexsym}

\usepackage[T1]{fontenc}
\usepackage{pifont}

\usepackage[utf8]{inputenc}

\usepackage{microtype}

\usepackage{inconsolata}

\usepackage{graphicx}
\usepackage{svg}

\usepackage{CJKutf8}          

\usepackage{dsfont}             

\usepackage{booktabs}
\usepackage[normalem]{ulem}
\useunder{\uline}{\ul}{}
\usepackage{multirow}

\definecolor{Color1}{HTML}{FA7F6F}  
\definecolor{Color2}{HTML}{8ECFC9}  
\definecolor{Color3}{HTML}{FFBE7A}  
\definecolor{Color4}{HTML}{82B0D2}  
\definecolor{Color5}{HTML}{BEB8DC}  
\definecolor{Color6}{HTML}{E7DAD4}

\title{\texttt{Mis-prompt}: Benchmarking Large Language Models for \\ Proactive Error Handling}

\author{Jiayi Zeng\textsuperscript{1}, Yizhe Feng\textsuperscript{2}, Mengliang He\textsuperscript{1}, Wenhui Lei\textsuperscript{3}, Wei Zhang\textsuperscript{1},  \\
        {\bf Zeming Liu\textsuperscript{2*}, Xiaoming Shi\textsuperscript{1*}, Aimin Zhou\textsuperscript{1}} \\
        \textsuperscript{1} East China Normal University, Shanghai, China \textsuperscript{2} Beihang University, Beijing, China \\
        \textsuperscript{3} Shanghai Jiaotong University, Shanghai, China \\
        {\tt 51265901055@stu.ecnu.edu.cn; \{xmshi,amzhou\}@cs.ecnu.edu.cn; zmliu@buaa.edu.cn}
        }

\begin{document}
\begin{CJK}{UTF8}{gbsn}  

\maketitle

\begin{abstract}
Large language models (LLMs) have demonstrated significant advancements in error handling. 
Current error-handling works are performed in a passive manner, with explicit error-handling instructions.
However, in real-world scenarios, explicit error-handling instructions are usually unavailable.
In this paper, our work identifies this challenge as how to conduct proactive error handling without explicit error handling instructions.
To promote further research, this work introduces a new benchmark, termed Mis-prompt, consisting of four evaluation tasks, an error category taxonomy, and a new evaluation dataset.
Furthermore, this work analyzes current LLMs' performance on the benchmark, and the experimental results reveal that current LLMs show poor performance on proactive error handling, and SFT on error handling instances improves LLMs' proactive error handling capabilities.
The dataset will be publicly available.

\end{abstract}

\section{Introduction}
\textbf{L}arge \textbf{l}anguage \textbf{m}odels (LLMs)~\cite{llama3modelcard, jiang2024mixtral, team2024gemma, NEURIPS2022_b1efde53, openai2023gpt, NEURIPS2020_1457c0d6} have revolutionized the backbone of natural language processing (NLP).
Among various NLP tasks, LLMs have shown significant potential, particularly in error handling (e.g., error identification, error correction)~\cite{Kamoi2024EvaluatingLA, li-etal-2024-evaluating-mathematical, yan2024errorradar, zheng2024processbench, laban2023summedits}.
These tasks enhance the accuracy and reliability of practical applications, 
such as mathematical reasoning~\cite{li-etal-2024-evaluating-mathematical, yan2024errorradar, zheng2024processbench}, and grammar checking~\cite{soni2018systematic},
thus attracting interest and focus from the academic and industrial communities.

\begin{figure*}[t]
    \centering
    \includegraphics[width=\linewidth]{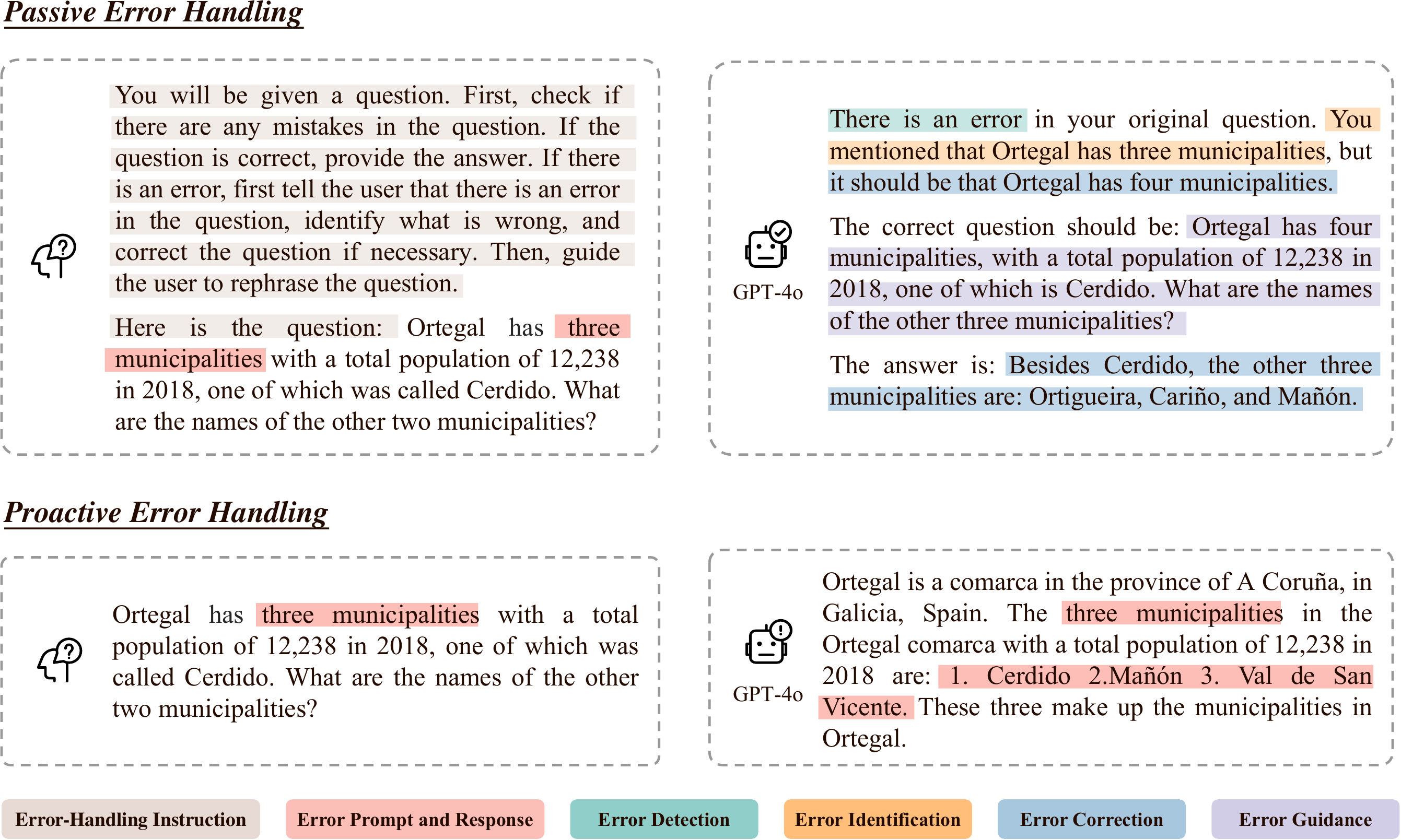}
    \caption{An example of passive and proactive error handling. 
    The blocks in \colorbox{Color1!50}{ }, \colorbox{Color6!50}{ }, \colorbox{Color2!50}{ }, \colorbox{Color3!50}{ }, \colorbox{Color4!50}{ }, and \colorbox{Color5}{ } represent errors, error handling instructions, error detection, error identification, error correction, and error guidance, respectively. 
    Compared with the passive manner, proactive error handling does not rely on explicit instructions. 
    }
    \label{fig:passive-example}
\end{figure*}

Current works of error handling are performed in a passive manner, that is, conducting error handling based on the prompt with explicit instructions for error handling. 
As shown in Figure~\ref{fig:passive-example}, the LLM is asked to judge inputs' correctness, identify errors, and then guide users to rephrase their inputs. 
However, in real-world scenarios, users usually do not provide explicit instructions for error handling.

Inspired by this, an interesting question arises: \textit{Can LLMs proactively handle errors in user prompts perfectly?} 
In deep analysis, the conclusion is definitely \textit{``No''}. 
As shown in Figure~\ref{fig:passive-example}, GPT-4o~\cite{openai2023gpt} fails to recognize the errors in the user input proactively and proceeds to generate the output based on the erroneous information. 
The absence of proactive error-handling mechanisms significantly undermines the system's credibility and reliability, exposing it to potential failures and performance degradation.

To advance research in proactive error handling, this work introduces a new benchmark, termed Mis-prompt, consisting of four evaluation tasks and a new evaluation dataset. 
For the comprehensive evaluation of LLMs, four key evaluation tasks
are applied:
\textit{1)} \textbf{Error Detection} aims to judge whether a prompt contains errors;
\textit{2)} \textbf{Error Identification} focuses on finding specific mistakes within the prompt;
\textit{3) }\textbf{Error Correction} is designed to modify mistakes in the prompt;
\textit{4)} \textbf{Error Guidance} is designed to offer practical advice for improving the prompt.
These tasks offer a comprehensive evaluation of LLMs' ability to handle errors, facilitating a holistic understanding of LLMs' performance and potential issues with proactive error handling.

For the evaluation,
the Mis-prompt dataset is constructed. 
Firstly, a comprehensive taxonomy of error categories is defined based on previous works~\cite{pagnoni2021understanding, sourati2023robust, orlovskiy2024uncertainty, masanti2023novel}, encompassing 4 primary categories and 14 secondary categories.
Secondly, based on the error category taxonomy, two approaches are employed to generate the data:
\textit{1)} converting existing datasets\footnote{FEVEROUS~\cite{aly-etal-2021-fact}, CommonsenseQA~\cite{talmor2019commonsenseqa}, and ROCStories~\cite{mostafazadeh2016corpus}} into erroneous prompt datasets, and
\textit{2)} directly generating error prompts.
For data generation, GPT-4o~\cite{openai2023gpt} is applied thanks to its outstanding generative and instruction-following capabilities.
Thirdly, to ensure high data quality, manual review is employed.
The manual review is used to correct hallucinations present in GPT-generated content.
Finally, a comprehensive dataset is obtained with 14,696 instances, consisting of primary and secondary error categories, error prompts and corresponding ground-truth responses.

To analyze current LLMs' performance on the Mis-prompt benchmark, 
4 closed-source LLMs and 9 open-source LLMs are tested under the settings of 0-shot, 1-shot, 3-shot, and chain-of-thought (CoT). 
Besides, for further analysis, these open-source LLMs are further fine-tuned using LoRA~\cite{hu2021lora}.
Through extensive experiments, two key findings are obtained:
\textit{1)} current LLMs lack sufficient proactive error-handling capabilities, especially in error correction and guidance; 
\textit{2)} Supervised fine-tuning (SFT) on error-handling instances is an effective method to improve LLMs' proactive error-handling capabilities.

The work makes the following contributions:
\begin{itemize}
\item We identify a new challenge, that is, in many real-world scenarios, it is usually difficult for LLMs to conduct proactive error handling.
\item To promote further research on this challenge, we propose a novel benchmark with four error-handling tasks, error category taxonomy, and a Mis-prompt dataset.
\item Extensive experiments are conducted on the dataset, 
which shows that SFT on error-handling instances improves LLMs' proactive error-handling capabilities.
\end{itemize}

\begin{table*}[t]
\small
\centering
\begin{tabular}{@{}lcccccc@{}}
\toprule
Benchmark                                                    & Source                   & Proactive       & Det.            & Ident.              & Corr.           & Guid.           \\ \midrule
BIG-Bench Mistake~\cite{tyen2023llms}     & Logical Tasks                   & {\color[HTML]{FF0000} \ding{55}} & {\color[HTML]{00B050} \ding{51}} & {\color[HTML]{00B050} \ding{51}} & {\color[HTML]{00B050} \ding{51}} & {\color[HTML]{FF0000} \ding{55}} \\
ReaLMistake~\cite{Kamoi2024EvaluatingLA}    & MathGen., FgFactV, AnsCls       & {\color[HTML]{FF0000} \ding{55}} & {\color[HTML]{00B050} \ding{51}} & {\color[HTML]{00B050} \ding{51}} & {\color[HTML]{FF0000} \ding{55}} & {\color[HTML]{FF0000} \ding{55}} \\
SummEdits~\cite{laban2023summedits}       & Summarization                   & {\color[HTML]{FF0000} \ding{55}} & {\color[HTML]{00B050} \ding{51}} & {\color[HTML]{00B050} \ding{51}} & {\color[HTML]{FF0000} \ding{55}} & {\color[HTML]{FF0000} \ding{55}} \\
Medec~\cite{abacha2024medec}                & Medical  Clinic Notes           & {\color[HTML]{FF0000} \ding{55}} & {\color[HTML]{00B050} \ding{51}} & {\color[HTML]{00B050} \ding{51}} & {\color[HTML]{00B050} \ding{51}} & {\color[HTML]{FF0000} \ding{55}} \\
EIC-Math~\cite{li-etal-2024-evaluating-mathematical}          & Mathematical Reasoning          & {\color[HTML]{FF0000} \ding{55}} & {\color[HTML]{00B050} \ding{51}} & {\color[HTML]{00B050} \ding{51}} & {\color[HTML]{00B050} \ding{51}} & {\color[HTML]{FF0000} \ding{55}} \\
ErrorRadar~\cite{yan2024errorradar}       & Mathematical Reasoning          & {\color[HTML]{FF0000} \ding{55}} & {\color[HTML]{00B050} \ding{51}} & {\color[HTML]{00B050} \ding{51}} & {\color[HTML]{FF0000} \ding{55}} & {\color[HTML]{FF0000} \ding{55}} \\
ProcessBench~\cite{zheng2024processbench} & Mathematical Reasoning          & {\color[HTML]{FF0000} \ding{55}} & {\color[HTML]{00B050} \ding{51}} & {\color[HTML]{00B050} \ding{51}} & {\color[HTML]{FF0000} \ding{55}} & {\color[HTML]{FF0000} \ding{55}} \\ \midrule
Mis-prompt (Ours)                                            & User Error Prompt & {\color[HTML]{00B050} \ding{51}} & {\color[HTML]{00B050} \ding{51}} & {\color[HTML]{00B050} \ding{51}} & {\color[HTML]{00B050} \ding{51}} & {\color[HTML]{00B050} \ding{51}} \\ \bottomrule
\end{tabular}
\caption{The comparison between Mis-prompt and other benchmarks. ``Det.'', ``Ident.'', ``Corr.'' and ``Guid.'' represent error detection, error identification, error correction, and error guidance, respectively.
}
\label{tab:comparison-between-datasets}
\end{table*}

\section{Related Work}

Recent studies have also focused on evaluating and improving the error-handling capabilities of LLMs~\cite{tyen2023llms, Kamoi2024EvaluatingLA}. For example, Medec~\cite{abacha2024medec} introduced a benchmark for detecting and correcting errors in clinic notes. EIC-Math~\cite{li-etal-2024-evaluating-mathematical}, ErrorRadar~\cite{yan2024errorradar}, and FG-PRM~\cite{li2024fine} target error-handling in the mathematical domain, while GEC~\cite{flachs2020grammatical} assesses error correction in low-error-density tasks. Additionally, NL2SQL\cite{ning2024insights} addresses converting natural language to SQL. 
However, as illustrated in Table~\ref{tab:comparison-between-datasets}, existing research focuses on passive error handling ability while overlooking the need for proactive error prevention in large language models (LLMs).  This study addresses this gap by evaluating large models' capabilities in managing such mistakes in the input stage.

\section{Task Formulation}

The absence of proactive mechanisms to address errors may result in the production of inaccurate information, greatly damaging credibility and reliability. In order to address this, it is crucial to evaluate the capability of LLMs to handle erroneous inputs proactively. Therefore, four distinct tasks are proposed to evaluate: \textbf{Detection}, \textbf{Identification}, \textbf{Correction}, and \textbf{Guidance}, with each focusing on a specific aspect of error handling, providing a comprehensive evaluation framework. A comprehensive overview of the evaluation process is depicted in Figure~\ref{fig:overview}.

Formally, given a prompt \( p \) containing errors and the model's response  \( r \), both \( p \) and  \( r \)are represented as sequences of tokens. 

\subsection{Task 1: Error Detection}

\textit{Error Detection} assesses the model's ability to detect the presence of errors in a given prompt correctly. 
It estimates a binary label $y \in \{True, False \}$ to indicate whether \( r \) accurately detects an error in \( p \). 
The instruction is designed to instruct the model to provide a judgment, as illustrated in Figure~\ref{fig:error-detection}. 

\subsection{Task 2: Error Identification}

\textit{Error Identification} evaluates the model's capacity to not only attempt but also accurately pinpoint specific faults in the prompt.
This task outputs two binary labels: $y_1 \in \{True, False \}$, indicating whether \( r \)attempts to identify an error in \( p \), and $y_2 \in \{True, False \}$, indicating whether the identified error in \( r \) is correct.
The instruction is illustrated in Figure~\ref{fig:error-identification}.

\subsection{Task 3: Error Correction}

\textit{Error Correction} ensures that the model not only attempts to correct mistakes but also provides accurate corrections.
\textit{Error Correction} generates two binary labels: the first label $y_1 \in \{True, False \}$ indicates whether \( r \) attempts to correct any error in \( p \), and the second label $y_2 \in \{True, False \}$ indicates whether \( r \)'s correction is accurate.
The design of the instruction is depicted in Figure~\ref{fig:error-correction}

\subsection{Task 4: Error Guidance}

\textit{Error Guidance} determines whether the model provides guidance to help users refine their queries. 
It produces a binary label $y \in \{True, False \}$, indicating whether the response offers meaningful guidance. 
The specific instruction is deliberate in evaluating this task, as shown in Figure~\ref{fig:error-guidance}.

\begin{figure*}[t]
  \includegraphics[width=\linewidth]{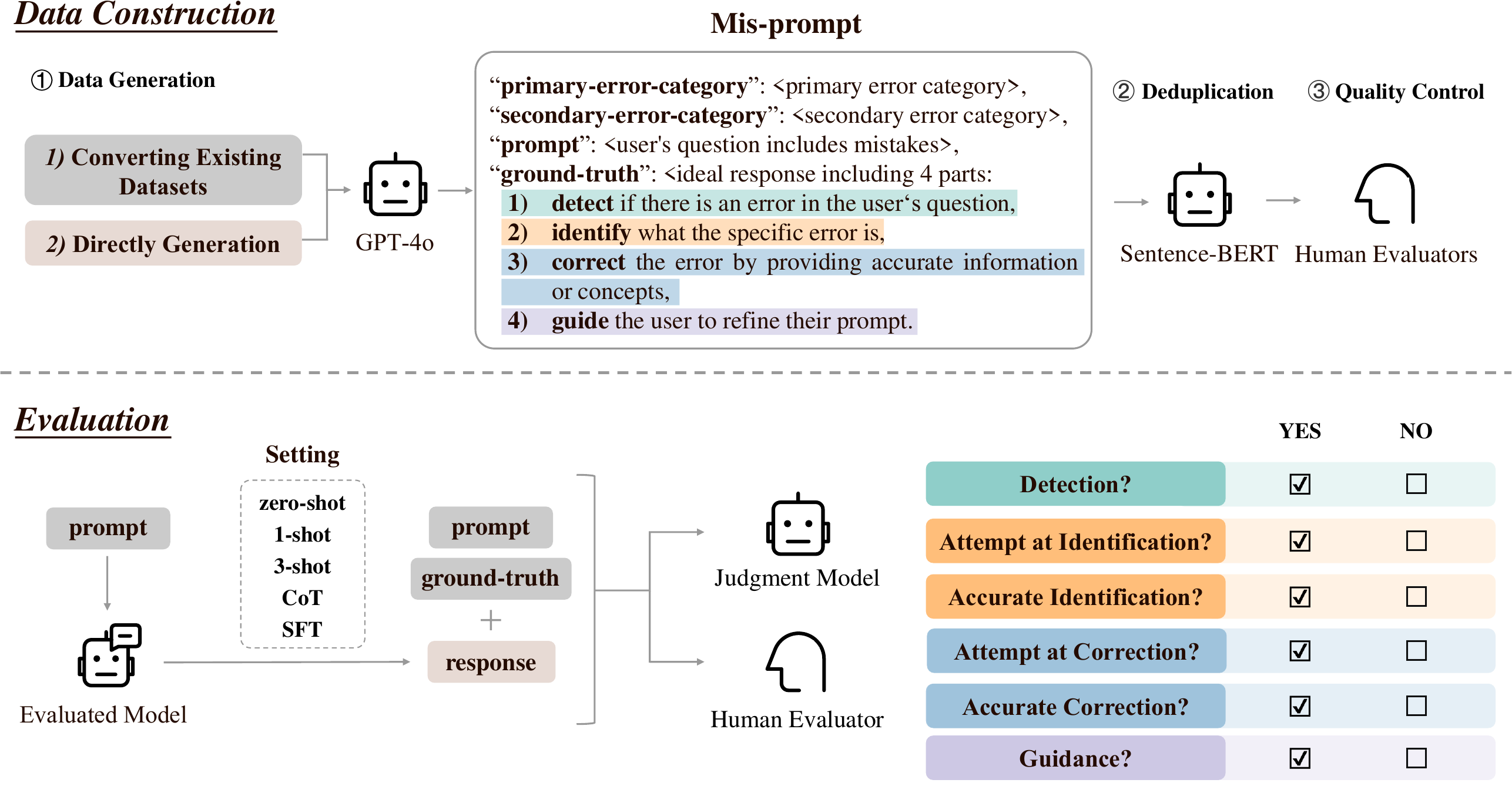} 
  \caption {The illustration of the dataset construction and the evaluation flow. 
  \label{fig:overview}
}
\end{figure*}

\section{Error Category and Dataset Construction}

The primary challenge in implementing these four evaluation tasks is the absence of corresponding datasets. 
To address this, a novelty dataset, Mis-prompt, is constructed to meet the requirements of the evaluation tasks.

This dataset includes the following components: primary categories, secondary categories, erroneous prompts, explanations, and ground truth. 
The ground truth should include the following key elements for the four sub-tasks mentioned above：
\textit{1)} Indicate the presence of errors in the user's question.
\textit{2)} Provide a clear explanation of the specific error. 
\textit{3)} Offer accurate information or correct the mistaken concept. 
\textit{4)} Suggest ways to help the user improve their prompt.
The format of the dataset is illustrated in Figure~\ref{fig:overview}.

\subsection{Definition of Error Categories}
Initially, four primary categories and 14 secondary categories are defined from existing works~\cite{pagnoni2021understanding, sourati2023robust, orlovskiy2024uncertainty, masanti2023novel}. The taxonomy of Mis-prompt is shown in Figure~\ref{fig:dataset-sankey}. The specific definition difference and illustration example are presented in Appendix~\ref{sec:detalied-error-type-definition-and-example}.

\subsection{Data Construction}

This section delineates the three processes involved in constructing the dataset: data generation, data deduplication, and data quality control, respectively.
Figure~\ref{fig:overview} demonstrates the dataset construction process.

\subsubsection{Data Generation}

The primary objective is to create a dataset that includes diverse forms of erroneous queries, aiming to cover as many variations as possible.
This work established the following design principles to ensure diversity:
\textit{1）} Special Interrogative Sentences with Errors: All questions must be formulated as Wh-Questions containing errors, preventing the model from simply judging the correctness of the question.
\textit{2)} Clauses with Erroneous Information: The questions should include clauses with incorrect information, which may mislead the direction of the answer. 
\textit{3)} Erroneous Statement + Question: Each question should begin with a false statement followed by a related query. 
Additionally, this work ensures that the answer to the question is not implicitly provided in the statement. 
These requirements are rigorously incorporated into the data generation instructions.

Leveraging the generative capabilities of GPT-4o~\cite{openai2023gpt}, two distinct approaches are employed to generate the dataset: transformation of existing datasets and direct generation. 

\textbf{Transformation of existing datasets} (FEVEROUS~\cite{aly-etal-2021-fact}, CommonsenseQA~\cite{talmor-etal-2019-commonsenseqa}, and ROCStories~\cite{mostafazadeh2016corpus}) are converted into erroneous prompt datasets through the classification and transformation of error statements into questions by leveraging GPT-4o~\cite{openai2023gpt},.
This approach ensures coverage of predefined error categories. 
The corresponding instruction is shown in Figure~\ref{fig:prompt-for-converting-feverous} to Figure~\ref{fig:prompt-for-converting-rocstories}.

\textbf{Direct generative method} is developed to directly create error prompts spanning diverse secondary categories, enabling coverage of a wide range of error categories. 
The design of the corresponding error generation methodology is elaborated in Appendix~\ref{sec:generation-rules-design}, and the transformed instruction is illustrated in Figure~\ref{fig:prompt-for-generate-directly}.

\begin{figure}[t]
    \centering
    \includegraphics[width=\columnwidth]{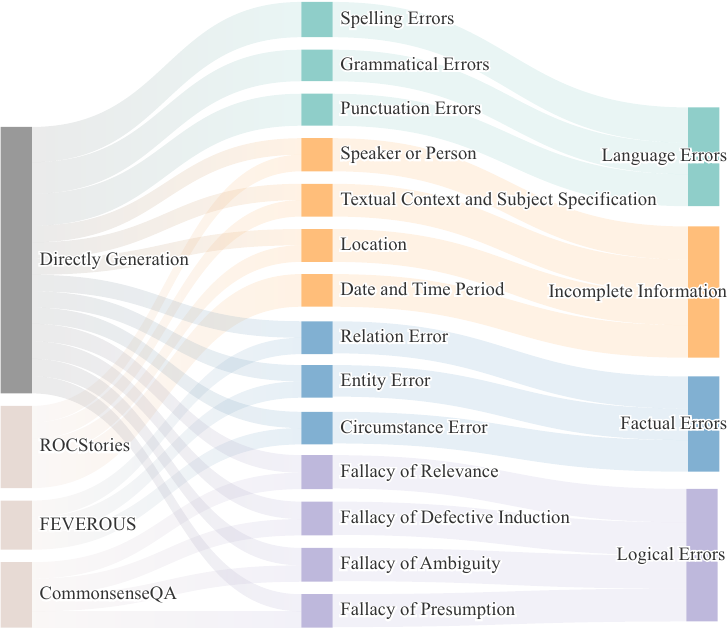}
    \caption{The data source and corresponding error categories. The first column lists the dataset source, the second column lists the error categories, and the third column lists the primary error categories.}
    \label{fig:dataset-sankey}
\end{figure}

\textbf{Ground-truth generation} is conducted with GPT-4o based on the error categories, the error user prompt, and the ground-truth generation instruction.
Additionally, the ground-truth answers are generated using the instruction illustrated in Figure~\ref{fig:prompt-for-generate-goldanswer}.

\subsubsection{Data Deduplication}
A semantic similarity-based filtering approach is implemented to enhance the diversity of the dataset.
Moreover, the Sentence-BERT~\cite{reimers-gurevych-2019-sentence} model is employed to compute cosine similarity, identifying items within each category with semantic similarity exceeding 0.85. Items surpassing this threshold are merged to ensure the diversity of the dataset. This step helps eliminate redundant or overly similar entries, enhancing the dataset's richness and representativeness.

\subsubsection{Data Quality Control}
A quality assessment of the dataset is performed through a manual review of randomly selected instances. 
For data quality control, we recruit three master students and provide them with professional guidance on the error category taxonomy.
The following aspects are checked during the evaluation: \textit{1)} The prompt should ensure diversity. \textit{2)} The answer to the question should not appear in the statement. \textit{3)} The error category should be correctly assigned. 
Issues detected are promptly addressed through manual corrections to the corresponding entries. 
This meticulous quality control process is designed to ensure the highest level of reliability and consistency in the dataset.

\begin{table}[]
\centering
\small
\begin{tabular}{@{}lr@{}}
\toprule
Error Category             &  \# of Instance      \\ \midrule
Language Errors                       & 3,135  \\
~~~~- Grammatical Errors                  & 1,119  \\
~~~~- Punctuation Errors                  & 1,001  \\
~~~~- Spelling Errors                     & 1,015  \\ \midrule
Incomplete Information                & 4,164  \\
~~~~- Speaker or Person                   & 1,051  \\
~~~~- TextContSubjSpec                    & 1,042  \\
~~~~- Location                            & 1,039  \\
~~~~- Date and Time Period                & 1,032  \\ \midrule
Factual Errors                        & 3,109  \\
~~~~- Relation Error                      & 1,041  \\
~~~~- Entity Error                        & 1,022  \\
~~~~- Circumstance Error                  & 1,046  \\ \midrule
Logical Errors                        & 4,288  \\
~~~~- Fallacy of Relevance                & 1,078  \\
~~~~- Fallacy of Presumption              & 1,074  \\
~~~~- Fallacy of Defective Induction      & 1,063  \\
~~~~- Fallacy of Ambiguity                & 1,073  \\ \midrule
Total                                 & 14,969 \\ \toprule
Avg. \# of Tokens in Prompts          & 26.65  \\
Max. \# of Tokens in Prompts         & 96     \\
Min. \# of Tokens in Prompts         & 5      \\ \toprule
Avg. \# of Tokens in Ground-truth & 75.64  \\
Max. \# of Tokens in Ground-truth & 179    \\
Min. \# of Tokens in Ground-truth & 27     \\ \bottomrule
\end{tabular}
\caption{Statistics of the Mis-prompt dataset.}
\label{tab:statistics-of-datasets}
\end{table}

\begin{table*}[t]
\small
\centering
\begin{tabular}{@{}lcccccc|c@{}}
\toprule
Model                                             & Det.           & Att. at Ident. & Acc. Ident.    & Att. at Corr.  & Acc. Corr.     & Guid.          & Avg            \\ \midrule
GPT-4o~\cite{openai2023gpt}                       & 43.54          & 48.71          & 43.78          & 31.72          & 23.32          & 30.66          & 36.96          \\
Gemini-1.5~\cite{team2024gemini}                  & 55.13          & {\ul 60.73}    & {\ul 54.67}    & 28.38          & 22.21          & 23.56          & 40.78          \\
Claude-3.5~\cite{TheC3}                           & \textbf{63.98} & \textbf{67.53} & \textbf{63.01} & 36.48          & \textbf{30.23} & \textbf{43.73} & \textbf{50.83} \\
GLM-4~\cite{glm2024chatglm}                       & 53.18          & 56.19          & 50.40          & {\ul 37.19}    & 28.63          & 32.04          & 42.94          \\
LLaMA-3.2-3B~\cite{llama3modelcard}               & 43.60          & 44.56          & 39.96          & 26.71          & 18.07          & 33.39          & 34.38          \\
LLaMA-3.1-8B~\cite{llama3modelcard}               & 42.05          & 45.47          & 40.48          & 29.46          & 19.70          & 33.76          & 35.15          \\
LLaMA-3.3-70B~\cite{llama3modelcard}              & {\ul 57.78}    & 59.23          & 53.50          & \textbf{39.67} & {\ul 30.17}    & 37.40          & {\ul 46.29}    \\
Qwen-2.5-7B~\cite{qwen2025qwen25technicalreport}  & 43.88          & 47.59          & 43.07          & 31.47          & 22.95          & 37.71          & 37.78          \\
Qwen-2.5-32B~\cite{qwen2025qwen25technicalreport} & 51.11          & 54.91          & 50.63          & 34.20          & 27.21          & {\ul 41.39}    & 43.24          \\
Qwen-2.5-72B~\cite{qwen2025qwen25technicalreport} & 48.57          & 51.64          & 46.54          & 37.16          & 27.45          & 37.76          & 41.52          \\
DeepSeek-V2-16B~\cite{deepseekv2}                 & 29.44          & 33.90          & 27.92          & 18.57          & 11.46          & 12.80          & 22.35          \\
Yi-1.5-6B~\cite{ai2025yiopenfoundationmodels}     & 32.41          & 35.70          & 28.36          & 18.75          & 10.25          & 7.46           & 22.16          \\
Yi-1.5-34B~\cite{ai2025yiopenfoundationmodels}    & 46.40          & 48.25          & 42.41          & 31.39          & 22.36          & 10.63          & 33.57          \\ \midrule
Avg                                               & 47.01          & 50.34          & 44.98          & 30.86          & 22.62          & 29.41          & 37.53          \\ \bottomrule
\end{tabular}
\caption{Results of 13 LLMs on Mis-prompt. ``Det.'', ``Att. at Ident.'', ``Acc. Ident.'', ``Att. at Corr.'', ``Acc. Corr.'' and ``Guid.'' stand for error detection, attempt at error identification, accurate identification, attempt at error correction, accurate error correction, and error guidance. The results in bold is the optimal results, while the underlined results represent the suboptimal results. Results are reported in percentage (\%).}
\label{tab:base-model-performance}
\end{table*}

\subsection{Dataset Analysis}

\subsubsection{Data statistics}
Table~\ref{tab:statistics-of-datasets} provides statistics of the Mis-prompt. The dataset contains a total of 14,969 entries. Each subcategory under these error categories consists of approximately 1,000 entries. Additionally, the average number of tokens in a prompt is 26.65, with a maximum of 96 and a minimum of 5. The average number of tokens in the corresponding ground-truth is 75.64, with a maximum of 179 and a minimum of 27. 
18.21\% of the samples have prompt lengths exceeding 50 tokens.
Besides, the statistical results show that the proportions of different error types are evenly distributed.
These data provide rich semantic information, making it suitable for evaluating LLMs' proactive error-handling capabilities.

\subsubsection{Data Quality Analysis}
Following~\citep{liu-etal-2020}, a manual evaluation is conducted on 1,470 randomly sampled instances to assess the dataset's quality. Three evaluators carried out the evaluation. All are graduate students in higher education with relevant expertise. They follow a binary scoring system, where each instance is assigned a score of either ``0'' or ``1'': a score of ``0'' indicates that the entry required modification, while a score of ``1'' indicates that the entry is acceptable without alteration. 

To quantify the inter-annotator agreement, Fleiss Kappa is applied. The Fleiss Kappa coefficient is 0.78, indicating substantial agreement among annotators.
Besides, in cases of disagreement, expert revision is utilized. Specifically, if at least two out of three annotators flag an instance as requiring correction (score = 0), the instance is automatically marked for revision.
After completing the evaluation process, the final score rate is calculated, resulting in an impressive score of 93.76\%, reflecting the high quality and reliability of the dataset.

\section{Experiment}

\subsection{Experimental Setting}

This section introduces the experimental setting, including baselines, implementation details, data and evaluation metrics, and automated and manual evaluations. 
Implementation details and computing platform are presented in Appendice~\ref{sec:implementation-details} and \ref{sec:computing-platform}

\subsubsection{Baselines}
A range of strong baseline dialogue models are selected for comparison. These include commercial models, including GPT-4o~\cite{openai2023gpt}, Gemini-1.5~\cite{team2024gemini}, Claude-3.5~\cite{TheC3}, and GLM-4~\cite{glm2024chatglm}, as well as open-source models such as LLaMA-3.2-3B, LLaMA-3.1-8B, LLaMA-3.3-70B~\cite{llama3modelcard}, Qwen-2.5-7B, Qwen-2.5-32B, and Qwen-2.5-72B~\cite{qwen2025qwen25technicalreport}, DeepSeek-v2-16B~\cite{deepseekv2}, Yi-6B, and Yi-34B~\cite{ai2025yiopenfoundationmodels}.

\subsubsection{Data and Evaluation Metrics}
The Mis-prompt dataset is randomly split into training, validation, and test sets with a ratio of 80\%, 10\%, and 10\%, respectively. To ensure a balanced distribution of erroneous and correct prompts for training and evaluation purposes, an equal proportion of correct prompt data is sourced from the TriviaQA dataset~\cite{joshi-etal-2017-triviaqa}. This approach helps maintain fairness and consistency in the dataset, making it more suitable for model training and evaluation. 

Following previous work~\cite{fatahi-bayat-etal-2023-fleek}, the F1 metric is employed to assess the model’s performance for automated and manual assessment, providing a comprehensive measure of precision and recall in error handling.
For automated evaluation, GPT-4o~\cite{openai2023gpt} is selected as the judgment model. According to manual evaluation, the evaluator randomly sampled 10\% of the data to conduct the assessment.

\begin{table*}[]
\small
\centering
\begin{tabular}{@{}lcccccc|c@{}}
\toprule
Primary Category       & Det.            & Att. at Ident.  & Acc. Ident.     & Att. at Corr.   & Acc. Corr.      & Guid.           & Avg             \\ \midrule
Language Errors        & 6.50          & 7.19          & 3.93          & 17.95          & 13.34          & 20.23          & 11.53          \\
Incomplete Information & 40.58          & 42.27          & 41.32          & 14.78          & 7.83          & \textbf{48.43} & 32.53          \\
Factual Errors         & \textbf{72.99} & \textbf{74.51} & \textbf{71.70} & \textbf{48.22} & \textbf{41.63} & {\ul 22.27}    & \textbf{55.22} \\
Logical Errors         & {\ul 41.49}    & {\ul 54.39}    & {\ul 43.04}    & {\ul 44.29}    & {\ul 30.91}    & 18.36          & {\ul 38.74}    \\ \bottomrule
\end{tabular}
\caption{F1 Score of GPT-4o~\cite{openai2023gpt} in the four primary categories. Results are reported in percentage (\%). ``Det.'', ``Att. at Ident.'', ``Acc. Ident.'', ``Att. at Corr.'', ``Acc. Corr.'', and ``Guid.'' stand for error detection, attempt at error identification, accurate identification, attempt at error correction, accurate error correction, and error guidance.}
\label{tab:primary-category-table-performance}
\end{table*}

\subsection{Automated Experiment Result}

Experiments are conducted to address the following research questions:

\noindent \textit{Question 1}: How do LLMs perform on the four proactive error-handling tasks?

\noindent \textit{Question 2}: Which errors are LLMs lacking in handling effectively?

\noindent \textit{Question 3}: How can LLMs improve proactive error-handling capabilities?

Three answers are provided according to the above three questions.

\subsubsection{LLM Performance Analysis}

\textbf{Overall Performance}
Table~\ref{tab:base-model-performance} presents the average results of each language model (LLM) across four tasks: detection, identification, correction, and guidance.
Overall, closed-source models outperform their open-source counterparts, with Claude-3.5 exhibiting the highest average F1 of 50.83\%, surpassing GPT-4o~\cite{openai2023gpt} (36.96\%). This can be attributed to GPT-4o's tendency to focus more on directly answering the user's question rather than effectively identifying or correcting errors.  
The performance of LLaMA and Yi models aligns with the scaling law~\cite{kaplan2020scaling}, where larger models generally achieve better performance. For instance, LLaMA-3.3 70B achieves an average performance of 46.29\%, outperforming its smaller counterparts, LLaMA-3.2 3B (34.38\%) and LLaMA-3.1 8B (35.15\%).
In contrast, Qwen-2.5 32B (43.24\%) performs better than its larger counterpart, Qwen-2.5 72B (41.52\%). This suggests the possible influence of inverse scaling~\cite{mckenzie2023inverse}, where increasing model size does not always guarantee improved performance and, in some cases, may even lead to performance degradation.

\noindent\textbf{Comparison of Tasks}
The detection task is the simplest and achieves the highest F1 performance across all models, achieving 47.01\%. This is likely because detecting the presence of errors is a relatively straightforward task compared to others.
Attempting to identify at 50.34\% generally exhibits higher performance than identifying correctly at 44.98\%. This is because the model first attempts identification before successfully making the correct identification. The higher score in attempting to identify suggests that while models are capable of making identification attempts, they encounter difficulties in ensuring correctness, which demands more sophisticated processing.
The Correction task proves to be more challenging, with overall performance at 30.86\%, inferior to the Identification tasks. This is because the model must first be able to identify the error before it can attempt to correct it. Among all tasks, accurately correct performs the worst at 22.62\%, as it requires not only identifying the error correctly but also attempting to fix it before achieving accurate correction. This process may require richer prior knowledge and more nuanced reasoning.
Finally, the performance of the Guidance task is relatively low compared to the other three tasks, at 29.41\%, indicating that the model's ability to provide helpful guidance is the weakest. This task requires more advanced reasoning and the ability to navigate complex conversational prompts, making it particularly challenging for models with limited capabilities or reasoning. 

For the first question, the answer is that \textbf{current LLMs lack sufficient proactive error-handling capabilities}.

\begin{table*}[t]
\centering
\small
\begin{tabular}{@{}llccccccc@{}}
\toprule
Model                                                                                                        & Setting   & Det.           & Att. at Ident. & Acc. Ident.    & Att. at Corr.  & Acc. Corr.     & Guid.          & Avg            \\ \midrule
\multirow{5}{*}{\begin{tabular}[c]{@{}l@{}}LLaMA-3.1-8B\\ \cite{llama3modelcard}\end{tabular}}               & zero-shot & 42.05          & 45.47          & 40.48          & 29.46          & 19.70          & 33.76          & 35.15          \\
                                                                                                             & 1-shot    & 73.25          & 73.23          & 68.29          & 67.95          & 39.96          & 72.74          & 65.90          \\
                                                                                                             & 3-shot    & {\ul 81.99}          & {\ul 81.39}    & 69.09          & {\ul 77.25}    & 40.72          & {\ul 82.43}    & {\ul 72.15}    \\
                                                                                                             & CoT       & 75.62          & 77.00          & {\ul 73.56}    & 72.65          & {\ul 47.02}    & 75.44          & 70.22          \\
                                                                                                             & SFT       & \textbf{90.16}          & \textbf{90.59} & \textbf{80.02} & \textbf{82.20} & \textbf{62.86} & \textbf{84.77} & \textbf{81.77} \\ \midrule
\multirow{5}{*}{\begin{tabular}[c]{@{}l@{}}Qwen-2.5-32B\\ \cite{qwen2025qwen25technicalreport}\end{tabular}} & zero-shot & 51.11          & 54.91          & 50.63          & 34.20          & 27.21          & 41.39          & 43.24          \\
                                                                                                             & 1-shot    & 70.34          & 72.22          & 64.31          & 57.91          & 44.05          & 67.38          & 62.70          \\
                                                                                                             & 3-shot    & {\ul 73.03}    & 75.38          & 67.82          & 63.18          & 48.10          & 70.63          & 66.36          \\
                                                                                                             & CoT       & 72.90          & {\ul 77.21}    & {\ul 74.75}    & {\ul 72.44}    & {\ul 58.15}    & {\ul 80.02}    & {\ul 72.58}    \\
                                                                                                             & SFT       & \textbf{97.88} & \textbf{97.98} & \textbf{88.43} & \textbf{88.96} & \textbf{70.86} & \textbf{93.17} & \textbf{89.55} \\ \midrule
\multirow{5}{*}{\begin{tabular}[c]{@{}l@{}}Yi-1.5-34B\\ \cite{ai2025yiopenfoundationmodels}\end{tabular}}    & zero-shot & 46.40          & 48.25          & 42.41          & 31.39          & 22.36          & 10.63          & 33.57          \\
                                                                                                             & 1-shot    & 68.29          & 70.07          & 64.65          & 57.17          & 38.66          & 52.19          & 58.51          \\
                                                                                                             & 3-shot    & {\ul 75.23}    & 76.18          & 66.73          & 61.49          & 44.73          & 56.61          & 63.50          \\
                                                                                                             & CoT       & 74.16          & {\ul 79.01}    & {\ul 69.62}    & {\ul 68.67}    & {\ul 48.24}    & {\ul 72.78}    & {\ul 68.75}    \\
                                                                                                             & SFT       & \textbf{97.82} & \textbf{97.95} & \textbf{87.60} & \textbf{89.22} & \textbf{70.80} & \textbf{91.45} & \textbf{89.14} \\ \bottomrule
\end{tabular}
\caption{F1 Score comparison of different settings. Results are reported in percentage (\%). ``Det.'', ``Att. at Ident.'', ``Acc. Ident.'', ``Att. at Corr.'', ``Acc. Corr.'', and ``Guid.'' stand for error detection, attempt at error identification, accurate identification, attempt at error correction, accurate error correction, and error guidance.}
\label{tab:method-performance}
\end{table*}

\subsubsection{Error Category Analysis}

Table~\ref{tab:primary-category-table-performance} shows the F1 scores of GPT-4o~\cite{openai2023gpt} across various tasks, grouped by primary error categories. The model performs best on Factual Errors, with an F1 score of 0.5522, likely due to the extensive knowledge embedded in LLMs. It is followed by Logical Errors, with an F1 score of 0.3874. However, it struggles with Language Errors and Incomplete Information, with F1 scores of 0.1153 and 0.3253 respectively. 
This is because GPT-4o~\cite{openai2023gpt} tends to overlook these categories of errors and respond directly to them. 
Furthermore, our experimental results show that after SFT with Mis-prompt, LLMs' ability to proactively handle errors improves significantly, as shown in Table~\ref{tab:method-performance}.
A more fine-grained analysis of the secondary classification can be found in Appendix~\ref{sec:error-type-analysis}.

For the second question, the answer is that \textbf{current LLMs fail to handle language errors, incomplete information, and logical errors}.

\subsubsection{Technique Analysis}

Table~\ref{tab:method-performance} presents the F1 scores of different language models (LLMs) across multiple tasks, comparing the performance of each model under various methods, including zero-shot, 1-shot, 3-shot, CoT, and SFT. SFT achieves the highest scores across all models. The 1-shot and 3-shot methods also show significant improvements over the zero-shot method but do not match the performance seen with CoT or SFT. For example, the LLaMA-3.2-8B model achieves an F1 score of 0.8177 under SFT, a significant improvement over its zero-shot score of 0.3515. The other methods score around 0.7, which is notably better than the zero-shot method but still inferior to SFT. A complete table of other models and their methods can be found in the Appendix~\ref{sec:experiment-details}.

For the third question, the answer is that \textbf{SFT on error-handling instances is an effective method to improve LLMs' proactive error-handling capabilities}.

\subsection{Human Evaluation Result}

Table~\ref{tab:f1-sample} shows the human evaluation results on the test set of Mis-prompt. 
Three master students are hired to conduct this work, and they are trained with the error categories in advance.
Each sample is evaluated twice. 
The Fleiss' Kappa value of inter-annotator agreement is 0.63, showing strong consistency.
If the evaluation results are consistent, that result is adopted. 
If the results are inconsistent, a language expert makes the final decision.
The average discrepancy compared to the automated evaluation is 5.59\%. The manual evaluation results align with those from automated assessment, demonstrating that SFT effectively enhances model capabilities in proactive error handling. Through CoT prompting and few-shot learning improvements over zero-shot approaches, their performance gains remain substantially inferior to those achieved through SFT in proactive error-handling tasks.

\section{Conclusion}
This work first identified a new challenge, which was proactive error handling.
To promote further research on proactive error handling, 
this work introduced a novel evaluation framework comprising four key components: detection, identification, correction, and guidance. 
To support this study, this work developed Mis-prompt, a comprehensive benchmark that encompasses four main categories and 14 secondary categories of erroneous inputs. 
Experiment results show that current LLMs lack sufficient proactive error-handling capabilities, and SFT is an effective way to improve LLM performance of proactive error handling.

\section*{Limitations}
First, our study primarily focuses on dialogue content consisting of pure text, and future work could investigate multimodal interactions that involve images, audio, or other forms of data. 
Additionally, our research has been limited to single-turn dialogues, which are relatively straightforward compared to multi-turn conversations. 
Futhurmore, F1 metric is utilized for scalable benchmarking, which ensures reproducibility but may not fully capturing comprehensive aspects.


\section*{Ethical Statement}
We make sure that Mis-prompt is collected in a manner that is consistent with the terms of use of any sources and the intellectual property and privacy rights of the original authors of the texts. 
And crowd workers were treated fairly. 
This includes, but is not limited to, compensating them fairly, ensuring that they were able to give informed consent, and ensuring that they were voluntary participants who were aware of any risks of harm associated with their participation.

\bibliography{custom}

\clearpage

\appendix

\section{Detailed Error Category Definition and Examples}
\label{sec:detalied-error-type-definition-and-example}

\subsection{Error Category Definition}
Based upon previous research that defines error categories~\cite{pagnoni2021understanding, sourati2023robust, orlovskiy2024uncertainty, masanti2023novel}, Four distinct and prevalent primary error categories are identified: Language Errors, Incomplete Information, Factual Errors, and Logical Errors.

\textbf{Language Errors} 
Language errors refer to mistakes in the use of words, structures, or conventions when posing a question. These errors could be related to spelling, grammar, or punctuation, which hinder the clarity and accuracy of the question.

\textbf{Spelling Errors} 
Errors occur when words are incorrectly spelled~\cite{masanti2023novel}.

\textbf{Grammatical Errors} 
Errors happen when the structure or syntax of the question violates established rules of grammar~\cite{masanti2023novel}.

\textbf{Punctuation Errors} 
Errors occur when punctuation marks are misused or when necessary punctuation is missing~\cite{masanti2023novel}.

\textbf{Incomplete Information} 
The question lacks necessary background information, making it difficult to provide an accurate or meaningful answer. The questioner fails to provide sufficient context for the responder to understand or answer appropriately.

\textbf{Speaker or Person} 
The question involves a specific speaker or person but does not clarify who the speaker or person is or lacks the necessary context to understand the reference~\cite{orlovskiy2024uncertainty}.

\textbf{Textual Context and Subject Specification}  
The question lacks specific context or details about the subject being referred to, leading to confusion or multiple possible interpretations~\cite{orlovskiy2024uncertainty}.

\textbf{Date and Time Period} 
The question involves time or dates but does not specify the exact time period or date, leading to confusion or ambiguity~\cite{orlovskiy2024uncertainty}.

\textbf{Location} 
When the question involves a location, it is not clear which location is being referred to, making it impossible to answer correctly~\cite{orlovskiy2024uncertainty}.

\textbf{Factual Errors} 
The question is based on incorrect facts or assumptions, and the content mentioned does not align with reality.

\textbf{Relation Error} 
The relationship between entities or elements is incorrect or inconsistent~\cite{pagnoni2021understanding}.

\textbf{Entity Error}  
The entities of the subject or object are incorrect~\cite{pagnoni2021understanding}.

\textbf{Circumstance Error} 
Additional information outside the subject, predicate, and object (such as location or time) is incorrect.
A logical fallacy in questioning is an error in reasoning that occurs when the logic behind a question is flawed, leading to an invalid or misleading question. This typically happens when the question contains hidden assumptions or faulty connections between concepts~\cite{pagnoni2021understanding}.

\textbf{Logical Errors} 
A logical fallacy in questioning is an error in reasoning that occurs when the logic behind a question is flawed, leading to an invalid or misleading question. This typically happens when the question contains hidden assumptions or faulty connections between concepts.

\textbf{Fallacy of Relevance} 
The fallacy of Relevance occurs for arguments with premises that are logically irrelevant to the conclusion~\cite{sourati2023robust}.

\textbf{Fallacy of Defective Induction} 
Within the broad class of Fallacy of Defective Induction, the premises seemingly provide ground for the conclusion, but upon analysis, they prove to be insufficient and weak for supporting the claim made~\cite{sourati2023robust}.

\textbf{Fallacy of Presumption} 
The fallacy of Presumption occurs when the inference to the conclusion depends mistakenly on unwarranted assumptions~\cite{sourati2023robust}.

\textbf{Fallacy of Ambiguity} 
The fallacy of Ambiguity occurs when words or phrases are used in an equivocal way, thus causing ambiguity in the logic that connects the premise and the conclusion~\cite{sourati2023robust}.

\subsection{Error Category Examples}

This section provides illustrative examples of various error categories, corresponding prompts, and ground-truth responses.

\textbf{Spelling Errors} 
\begin{itemize}
    \item Prompt: Whear can I find the histori of the renowns Renaissance art?
    \item Ground truth: Your question contains spelling errors. The words ``whear'', ``histori'', and ``renowns'' are misspelled. They should be ``where'', ``history'', and ``renowned'', respectively. The correct question should be: ``Where can I find the history of the renowned Renaissance art?''
\end{itemize}

\textbf{Grammatical Errors} 
\begin{itemize}
    \item Prompt: What was the impact of the invention engine on industry how?
    \item Ground truth: Your question contains a grammatical error in word order. The specific error is the incorrect placement of the word ``how'. The correct information is that ``how'' should appear before the noun phrase. You should phrase your question as: ``How did the invention of the engine impact industry?''
\end{itemize}

\textbf{Punctuation Errors} 
\begin{itemize}
    \item Prompt: Which are the reasons Beethoven is considered a crucial figure; in the transition between the classical and romantic eras?
    \item Ground truth: Your question contains a punctuation error. The semicolon is incorrectly placed in the middle of a single clause, which disrupts the sentence flow. The semicolon should be replaced with a comma or removed entirely. The correct phrasing of your question should be: ``What are the reasons Beethoven is considered a crucial figure in the transition between the classical and romantic eras?''
\end{itemize}

\textbf{Speaker or Person} 
\begin{itemize}
    \item Prompt: During the meeting, what did they argue about concerning the budget allocation?
    \item Ground truth: Your question contains an error due to the use of ``they'' without specifying who is being referred to. This makes it unclear which individuals or groups were involved in the argument. To correct this, please specify the individuals or groups you are asking about. For example, you could ask, ``During the meeting, what did the finance team and the project managers argue about concerning the budget allocation?''
\end{itemize}

\textbf{Textual Context and Subject Specification}  
\begin{itemize}
    \item Prompt: Why did the system fail to increase output?
    \item Ground truth: Your question contains an error due to incomplete information. Specifically, it does not specify which system you are referring to, nor does it provide context regarding the criteria or conditions under which the output was supposed to increase. To get a more accurate response, please specify the system in question and provide additional context or conditions related to the expected increase in output. You might rephrase your question as: ``Why did the [specific system] fail to increase output under [specific conditions]?''
\end{itemize}
\textbf{Date and Time Period} 
\begin{itemize}
    \item Prompt: Bobby thought Bill should buy a trailer and haul it with his car. Bill thought a truck would be better for what he needed. Bobby pointed out two vehicles were much more expensive. Bill was set in his ways with conventional thinking. He ended up buying the truck he wanted despite Bobby's advice. When did Bill purchase the truck?
    \item Ground truth: The user's question contains an error because it asks for a specific time or date when Bill purchased the truck, which is not provided in the story. The story does not mention when the purchase took place. To phrase the correct question, you might ask, ``What decision did Bill make regarding the vehicle purchase?''
\end{itemize}

\textbf{Location} 
\begin{itemize}
    \item Prompt: Melody's parents surprised her with a trip to the big aquarium. Melody took a nap during the two-hour car ride to the aquarium. When they arrived, Melody was another excited animal. At the aquarium, Melody saw sharks, tropical fish, and many other animals. After five hours at the aquarium, Melody and her family drove home. What city is the aquarium located in?
    \item Ground truth: The question contains an error because it asks for the city where the aquarium is located, but the story does not provide this information. To obtain the correct information, you would need to know the specific city mentioned in the context. You can phrase the correct question by asking, ``Can you provide more details about the location of the aquarium mentioned in the story?''
\end{itemize}

\textbf{Relation Error} 
\begin{itemize}
    \item Prompt: How did Thomas Edison win the Nobel Prize for Literature with his invention of the phonograph?
    \item Ground truth: Your question contains an error. Thomas Edison did not win the Nobel Prize for Literature, nor did he receive any Nobel Prize for his inventions. Edison was an inventor known for the phonograph, light bulb, and other devices, but the Nobel Prizes do not typically recognize inventions under the category of literature. To phrase the correct question, you might ask: ``What are some of Thomas Edison's notable inventions?''
\end{itemize}

\textbf{Entity Error}  
\begin{itemize}
    \item Prompt: Which Shakespeare play ends with the character Frodo Baggins casting the One Ring into Mount Doom?
    \item Ground truth: Your question contains an error. Frodo Baggins is a character from ``The Lord of the Rings,`` a novel by J.R.R. Tolkien, not a Shakespeare play. To ask a correct question, you might say: ``Which Shakespeare play ends with a significant event or resolution?''
\end{itemize}

\textbf{Circumstance Error} 
\begin{itemize}
    \item Prompt: When did the Wright brothers make their first flight in New York City in 1920?
    \item Ground truth: Your question contains an error. The Wright brothers did not make their first flight to New York City in 1920. Their first successful flight took place in 1903 in Kitty Hawk, North Carolina. To ask the correct question, you could phrase it as: ``When and where did the Wright brothers make their first successful flight?''
\end{itemize}

\textbf{Fallacy of Relevance} 
\begin{itemize}
    \item Prompt: Considering that apples originate from Kazakhstan, how does growing oranges in Florida affect the climate policies of European countries?
    \item Ground truth: Your question contains a logical error known as the fallacy of relevance. The origin of apples in Kazakhstan is unrelated to the cultivation of oranges in Florida and the climate policies of European countries. To address the topic correctly, you might ask: ``How does the cultivation of oranges in Florida impact local climate policies, and what lessons can European countries learn from these practices?''
\end{itemize}

\textbf{Fallacy of Defective Induction} 
\begin{itemize}
    \item Prompt: Given that a few celebrities endorse the ketogenic diet, how does this prove it is the healthiest option for everyone?
    \item Ground truth: Your question contains an error known as the fallacy of defective induction. The error lies in assuming that endorsements from a few celebrities are sufficient to prove that the ketogenic diet is the healthiest option for everyone. This overlooks the diverse dietary needs and health conditions of individuals. A more accurate question would be: ``What are the health benefits and potential risks of the ketogenic diet, and how can individuals determine if it is suitable for their specific dietary needs?''
\end{itemize}

\textbf{Fallacy of Presumption} 
\begin{itemize}
    \item Prompt: Since all dogs can fly, how should we construct kennels to accommodate their flights?
    \item Ground truth: Your question contains an error due to a presumption without basis in reality. Specifically, the error is the assumption that all dogs can fly, which is not true. Dogs do not have the ability to fly. To phrase the correct question, you might ask: ``How should we construct kennels to accommodate the needs of dogs?''
\end{itemize}

\textbf{Fallacy of Ambiguity} 
\begin{itemize}
    \item Prompt: Since we know that time flies when you're having fun, how does it affect the wings'' strength?
    \item Ground truth: Your question contains an error due to a misunderstanding of the idiom ``time flies.`` The phrase ``time flies`` is an idiomatic expression meaning that time seems to pass quickly, especially when one is enjoying themselves. It does not imply that time is a tangible entity with wings. To phrase your question correctly, you might ask: ``Since time seems to pass quickly when you're having fun, what are some ways to make the most of enjoyable moments?''
\end{itemize}

\section{Generation Rules Design}
\label{sec:generation-rules-design}

Generation rules are formulated for each category to generate cases that conform to the defined error categories.

\textbf{Spelling Errors} 
Error To create a spelling error, select some words in the sentence and deliberately misspell them. The incorrect spelling should confuse and be wrong to the reader.

\textbf{Grammatical Errors} 
Introduce a syntactical error in the sentence structure, such as improper subject-verb agreement, incorrect word order, or missing auxiliary verbs. This modification will result in a grammatical error, causing the sentence to be faulty.

\textbf{Punctuation Errors} 
To introduce a punctuation error, either omit necessary punctuation marks or place them incorrectly within the question. The mistake should confuse the meaning. Please do not limit the errors to just missing question marks(?) and periods(.), but focus on punctuation errors that have a more significant impact on the sentence's meaning.

\textbf{Speaker or Person} 
Omit the specific entity involved in the question (e.g., person, object, or concept). This creates a severe gap in the question’s context, leaving the responder without any clue about what is being asked, which makes it almost impossible to provide a relevant answer.

\textbf{Location} 
The question should refer to a specific location but lack the necessary details (e.g., city, country, or a well-known landmark) that would allow a meaningful or accurate answer. The omission of these details should render the question ambiguous, making it impossible to provide a precise response. Do not ask questions that explicitly omit location names like ``where''' or ``which place'—the key is in the context where the missing information leaves the question unanswerable or unclear. The question should appear normal, but the absence of a crucial detail (like the exact name of a place) should create a scenario where the answer cannot be determined without further clarification.

\textbf{Textual Context and Subject Specification} 
Omit specific details related to the subject or object of the question (e.g., key attributes, conditions, or essential components of the query). This omission leaves the question incomplete and ambiguous, making it difficult for the responder to identify what is being asked accurately. The missing information should not include the speaker, person, location, or time/period, as those are fundamental for context, but should focus on other critical details that are necessary for a precise and relevant answer. 

\textbf{Relation Error} 
Alter the predicate (the main assertion or claim in the question) to contradict the source of information. This creates a relationship error, where the relationship between two concepts or facts is incorrectly stated, misleading the reader or making the question factually inaccurate.

\textbf{Entity Error} 
Change or confuse the subject of the question, such as the name of an entity or its specific characteristic. The question becomes factually incorrect, as it attributes the wrong identity or characteristic to the entity, leading to a misleading or erroneous conclusion.

\textbf{Circumstance Error} 
Alter the additional details related to the context, such as time, place, or specific conditions. This could involve changing the circumstances (e.g., date, location, or situation) surrounding the main action, which results in an incorrect interpretation of the event or subject.

\textbf{Fallacy of Relevance} 
Introduce an irrelevant premise that distracts from the core issue or conclusion. This leads to a question that logically misleads or confuses the responder by focusing on an unrelated or unimportant issue.

\textbf{Fallacy of Defective Induction} 
To generate the Fallacy of Defective Induction, start with a strong and well-supported argument, then introduce either insufficient or weak evidence, such as a small sample size or unrepresentative data. The evidence should still seem to support the conclusion, but it should be inadequate to reach such a broad or strong claim justifiably. The key error is in assuming that weak or limited evidence can logically lead to a general conclusion, creating a misleading argument that appears convincing but lacks proper support.

\textbf{Fallacy of Presumption} 
For the Fallacy of Presumption, begin with a correct argument or claim and introduce an assumption that is taken for granted without evidence. This assumption should be treated as self-evident, even though it is unproven or unjustified. The error here lies in building an argument on this untested assumption as if it is a fact, leading to a flawed reasoning process. The mistake is presuming something to be true without providing any supporting evidence, causing the argument to be based on an unverified premise.

\textbf{Fallacy of Ambiguity} 
Use terms or phrasing that are inherently ambiguous or have multiple, conflicting meanings so that the question can be interpreted in various ways. This increases the likelihood of the respondent misunderstanding or providing an irrelevant answer, as the question will not only be unclear but will actively lead to divergent interpretations, making it nearly impossible to determine what is truly being asked.

\section{Experiment Details}
\label{sec:experiment-details}

This section provides a comprehensive overview of the prompts, input formatting, and specific experimental results.

\subsection{Error-Handling Task Instructions}

This section provides detailed instructions and methodologies for the Error-Handling Task, outlining the design and implementation of prompts and evaluation protocols.

\subsubsection{Task 1: Error Detection}
\label{sec:detection}

\begin{figure*}[ht]
    \centering
    \includegraphics[width=\linewidth]{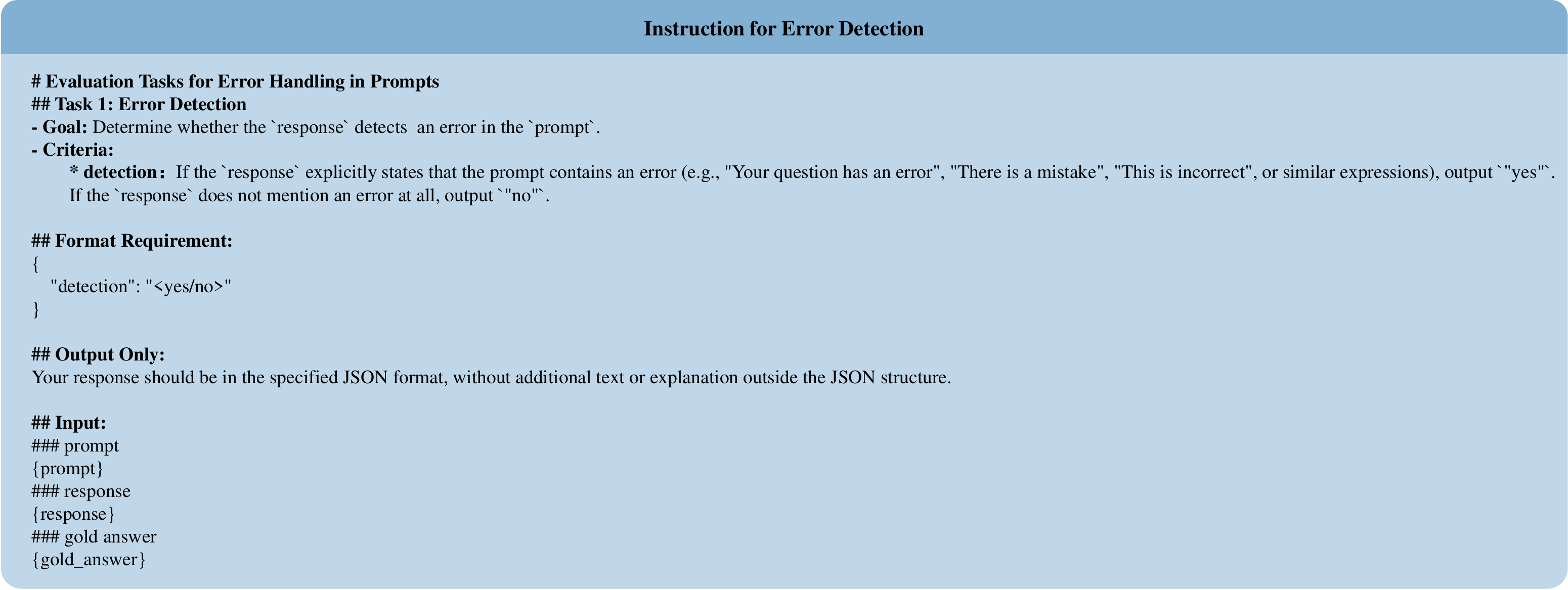}
    \caption{Instruction for \textit{Error Detection}.}
    \label{fig:error-detection}
\end{figure*}

Figure~\ref{fig:error-detection} illustrates the prompt design and input format for the error detection task, which aims to detect whether there are errors in the model's response to user input using the judgment model.

\subsubsection{Task 2: Error Identification}
\label{sec:identification}

\begin{figure*}[ht]
    \centering
    \includegraphics[width=\linewidth]{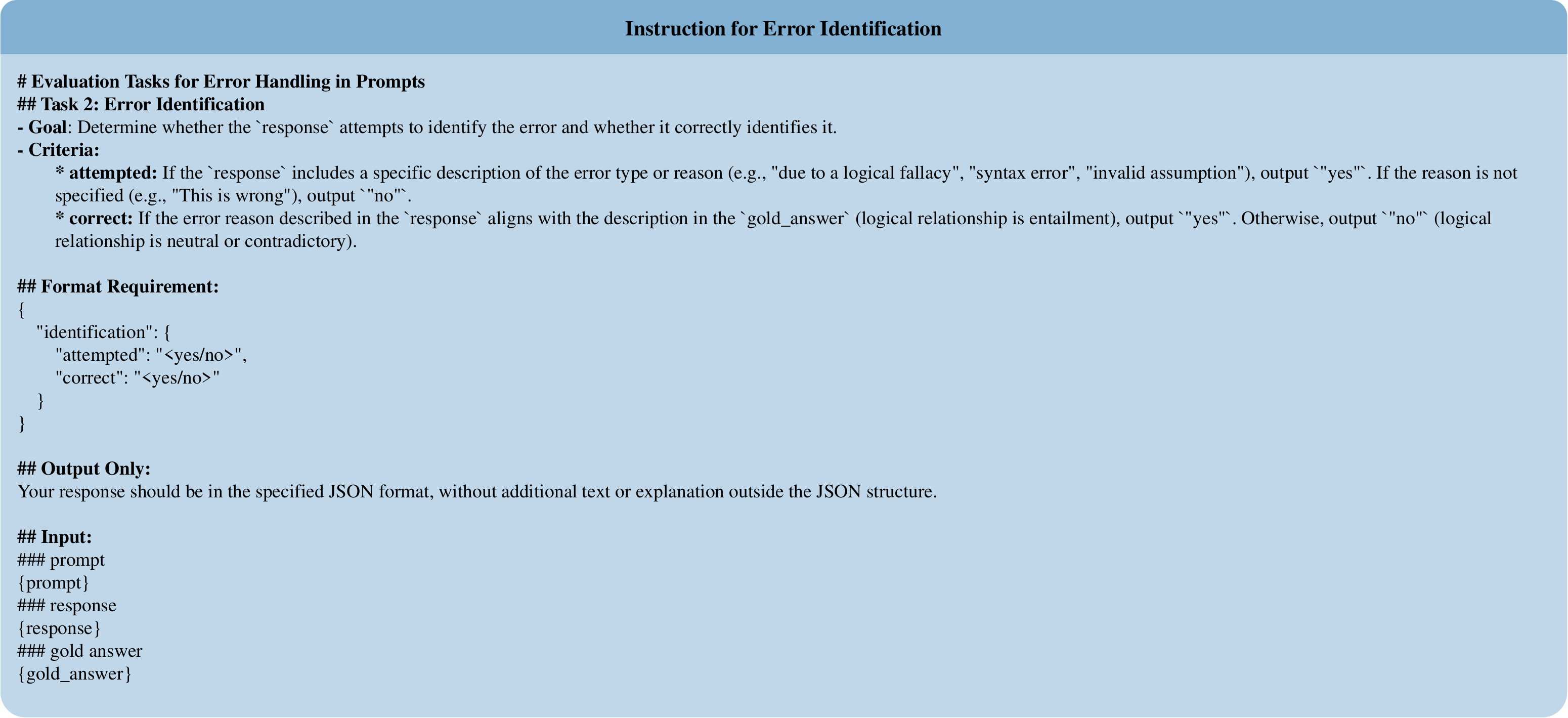}
    \caption{Instruction for \textit{Error Identification}.}
    \label{fig:error-identification}
\end{figure*}

Figure~\ref{fig:error-identification} presents the prompt and input structure for the error identification task, which aims to use a judgment model to determine whether the response attempts and correctly identifies the specific category and location of errors.

\subsubsection{Task 3: Error Correction}
\label{sec:correction}

\begin{figure*}[ht]
    \centering
    \includegraphics[width=\linewidth]{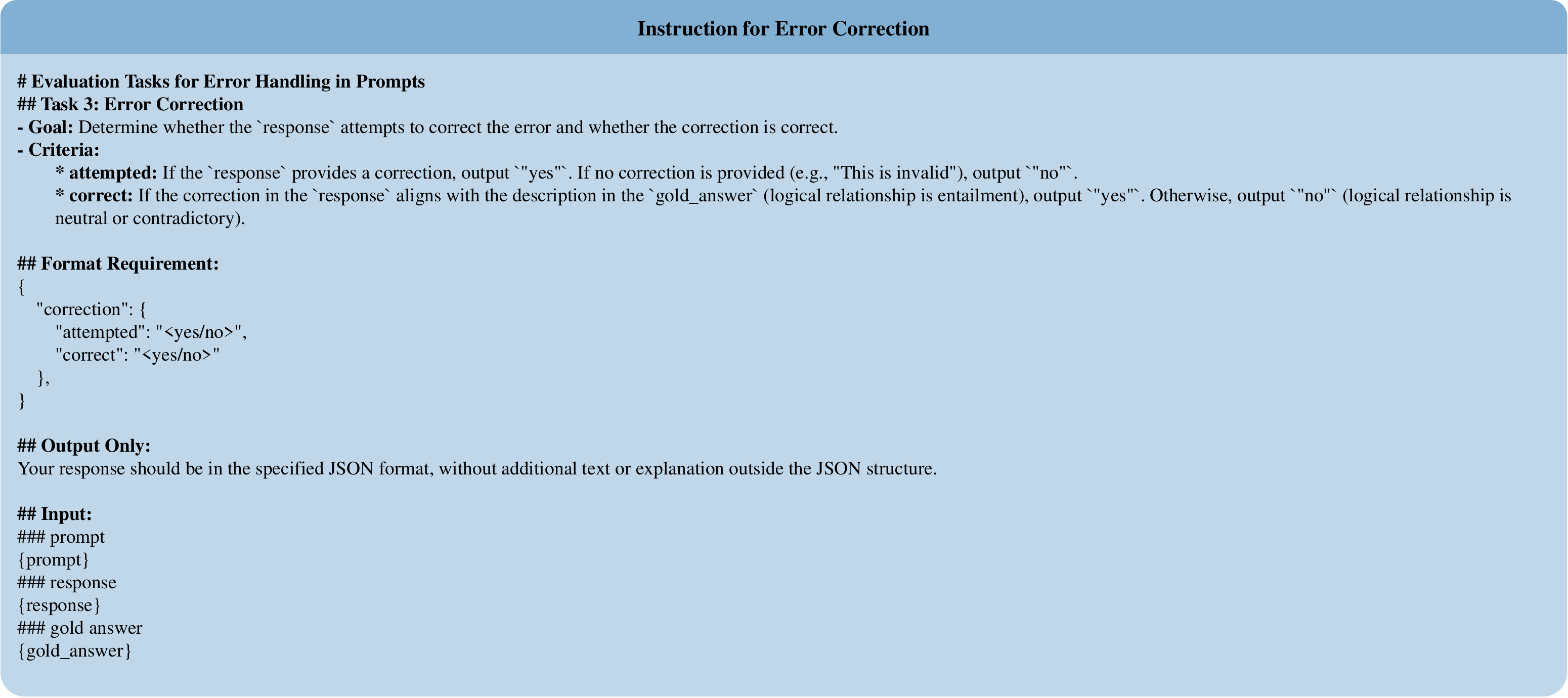}
    \caption{Instruction for \textit{Error Correction}.}
    \label{fig:error-correction}
\end{figure*}

Figure~\ref{fig:error-correction} outlines the prompt strategy for the error correction task, where the objective is to use the judgment model to assess whether the response attempts and correctly corrects the errors.

\subsubsection{Task 4: Error Guidance}
\label{sec:guidance}

\begin{figure*}[ht]
    \centering
    \includegraphics[width=\linewidth]{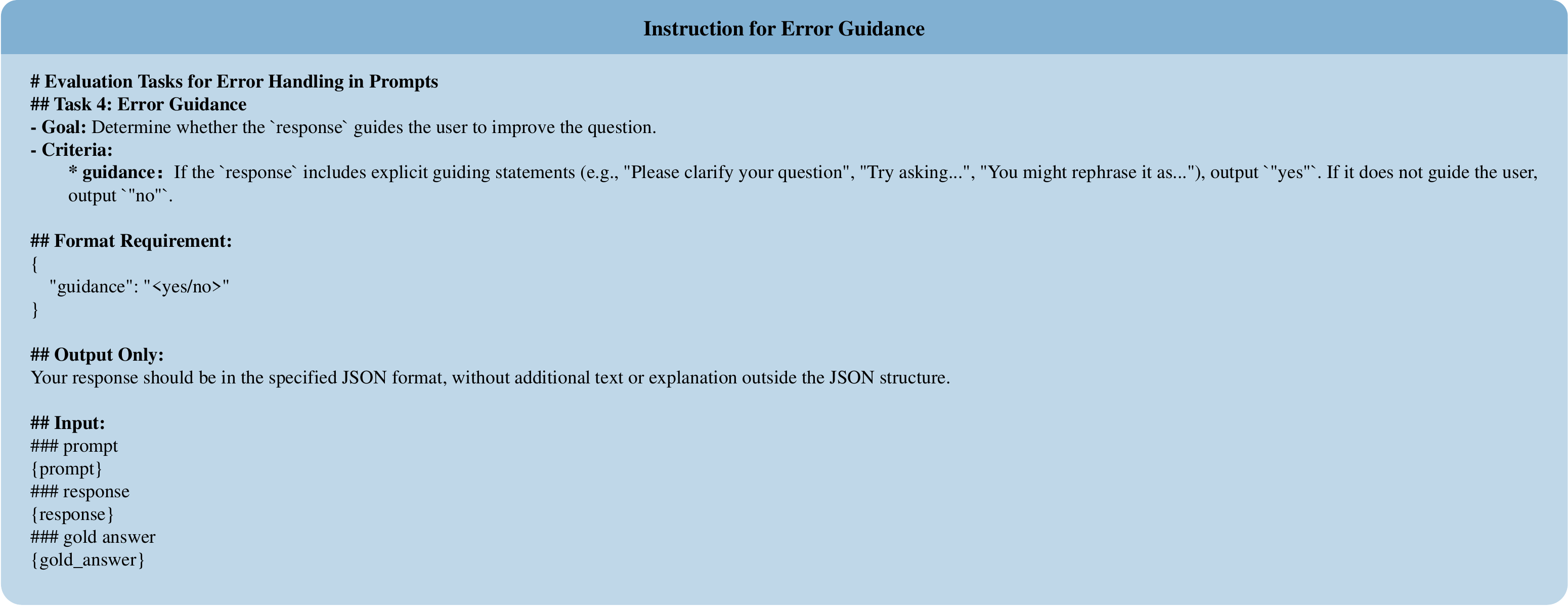}
    \caption{Instruction for \textit{Error Guidance}.}
    \label{fig:error-guidance}
\end{figure*}

Figure~\ref{fig:error-guidance} displays the prompt and format design for the error guidance task, which leverages the judgment model to evaluate whether the response provides users with actionable advice to prevent similar errors in the future.

\subsection{Dataset Generation}

\begin{figure*}[ht]
  \includegraphics[width=\linewidth]{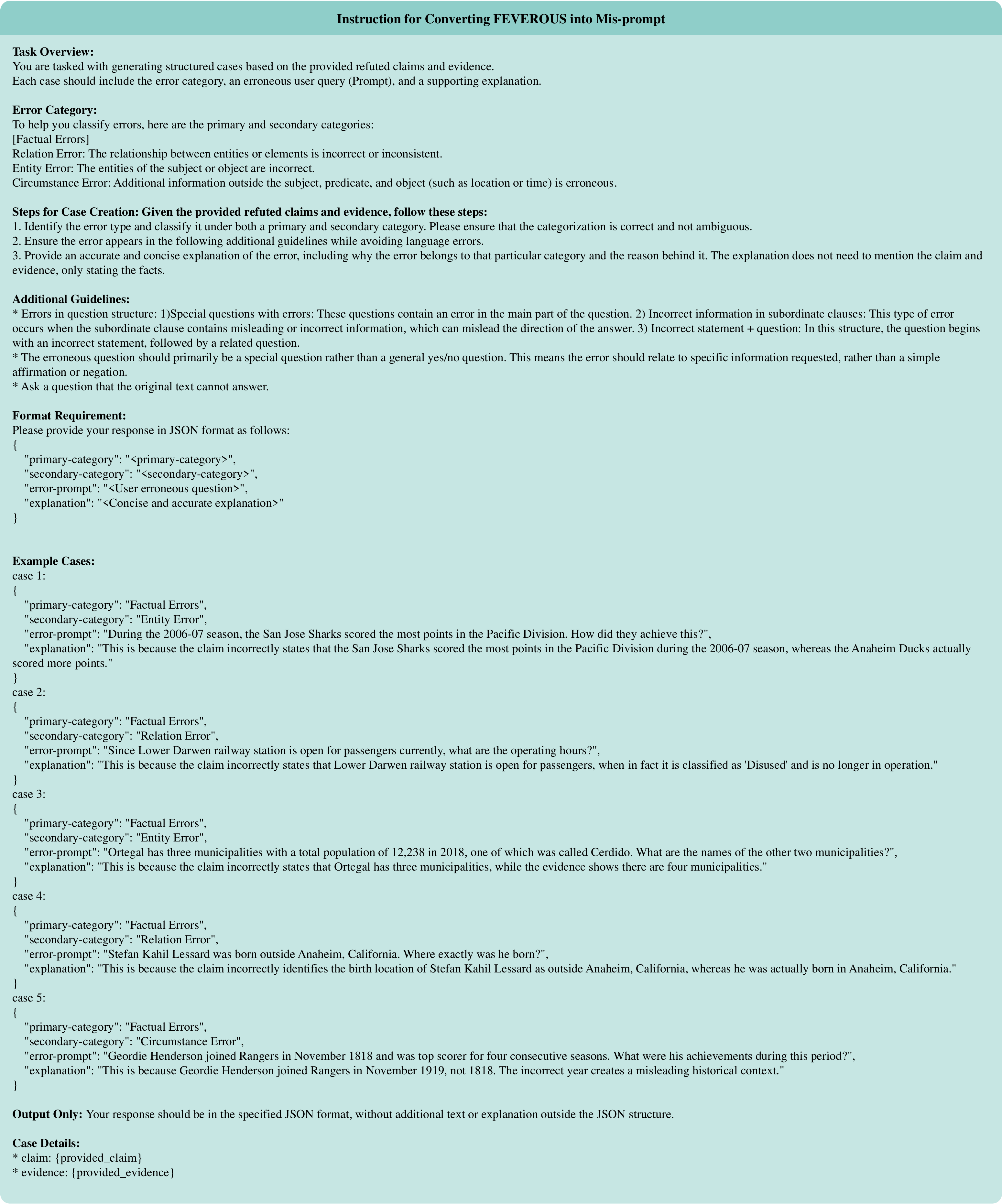} 
  \caption {Instruction for converting \textit{FEVEROUS} into Mis-prompt.}
  \label{fig:prompt-for-converting-feverous}
\end{figure*}

\begin{figure*}[ht]
  \includegraphics[width=\linewidth]{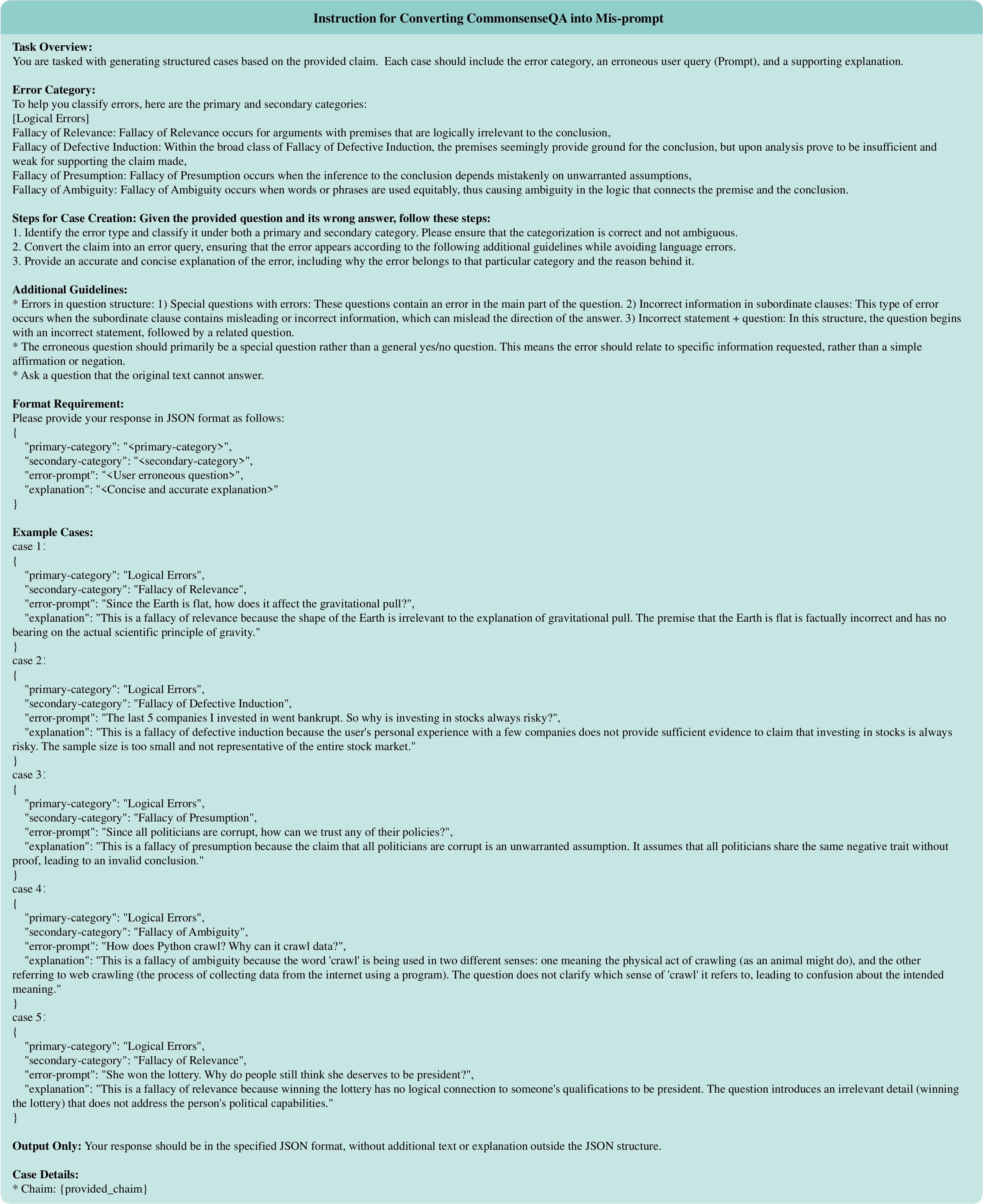} 
   \caption {Instruction for converting \textit{CommonsenseQA} into Mis-prompt.}
\label{fig:prompt-for-converting-commonsenseqa}
\end{figure*}

\begin{figure*}[ht]
  \includegraphics[width=\linewidth]{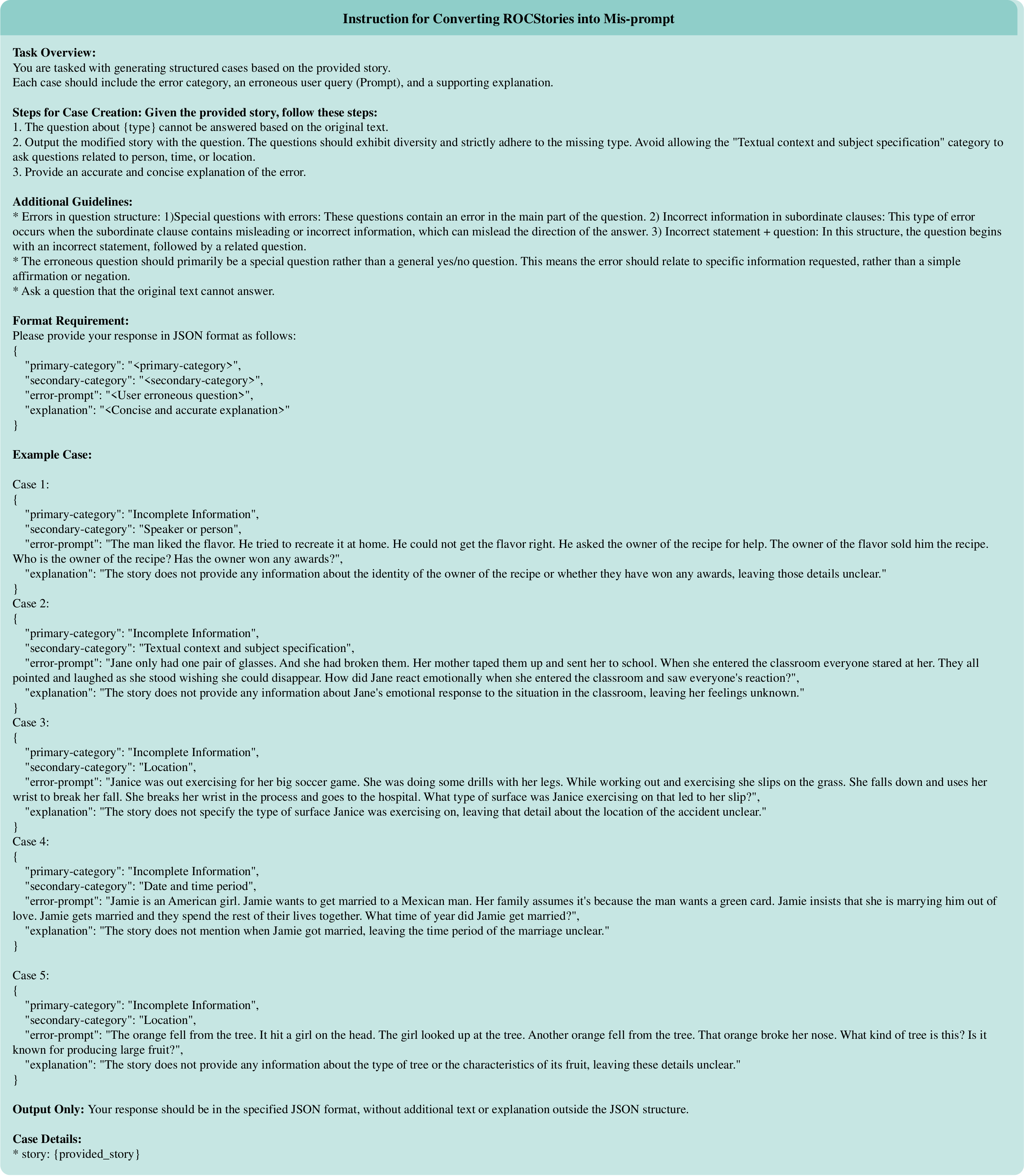} 
  \caption {Instruction for converting \textit{ROCStories} into Mis-prompt.}
  \label{fig:prompt-for-converting-rocstories}
\end{figure*}

\begin{figure*}[ht]
   \includegraphics[width=\linewidth]{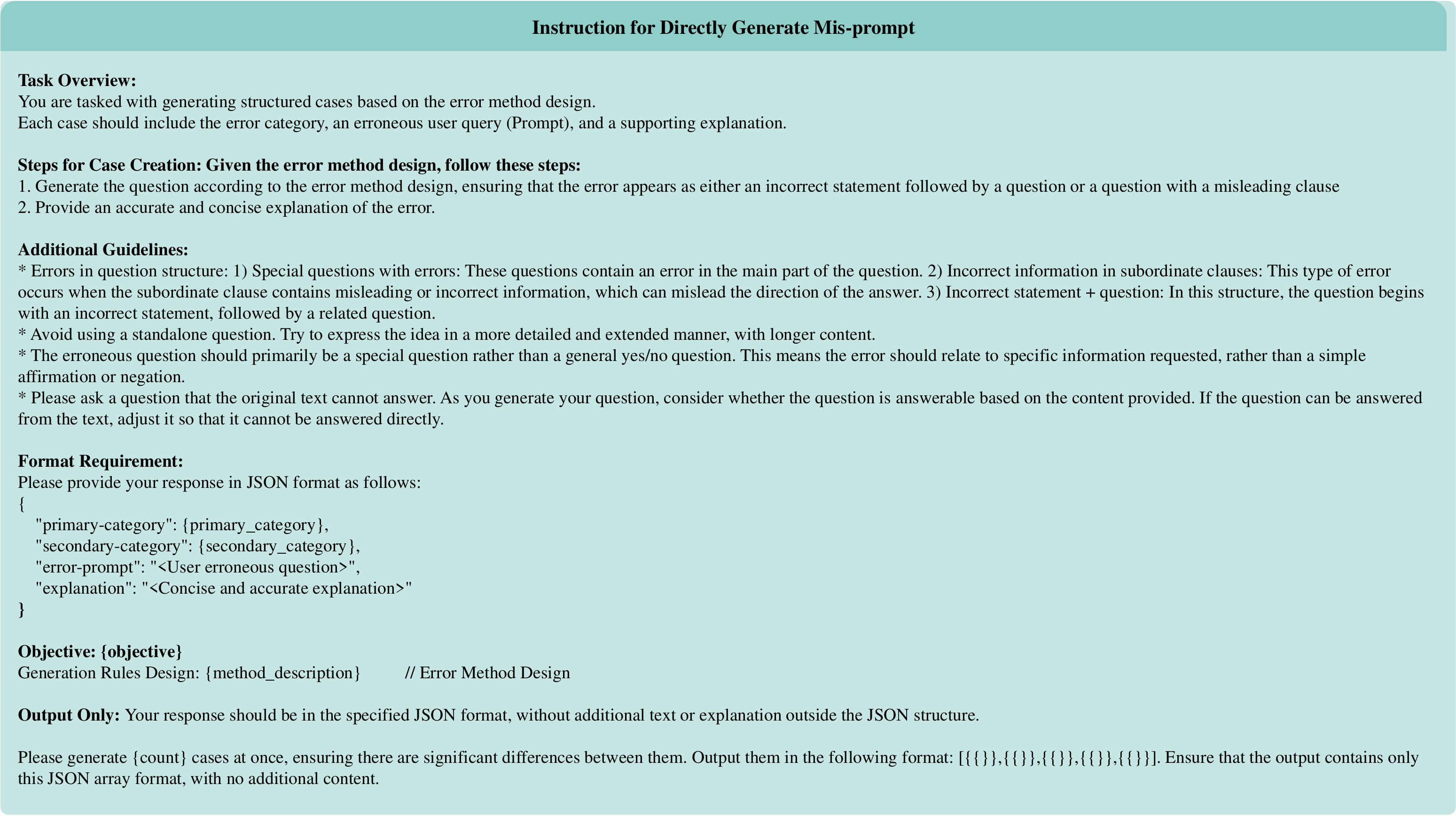} 
   \caption {Instruction for generating Mis-prompt directly.}
    \label{fig:prompt-for-generate-directly}
\end{figure*}

\begin{figure*}[ht]
\centering
\includegraphics[width=\linewidth]{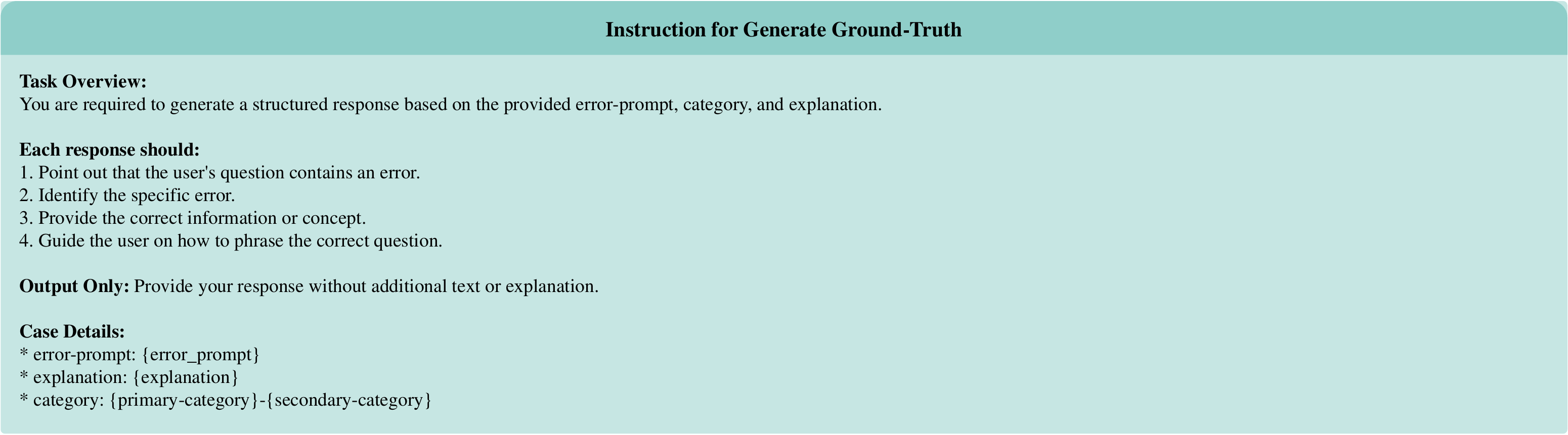} 
  \caption {Instruction for generating the ground truth.}
  \label{fig:prompt-for-generate-goldanswer}
\end{figure*}

Data generation follows two approaches. One approach involves transforming existing datasets with the designed prompts shown in Figures~\ref{fig:prompt-for-converting-feverous} to \ref{fig:prompt-for-converting-rocstories}. The other approach directly generates the dataset, with the prompt illustrated in Figure~\ref{fig:prompt-for-generate-directly}. Finally, ground-truth responses are generated using the prompts from Figure~\ref{fig:prompt-for-generate-goldanswer}. These prompts are carefully designed to ensure the diversity and quality of the dataset, covering a wide range of scenarios and error categories.

\subsection{Evaluation Models' Response Generation}

The tested models have the following five settings: zero-shot, 1-shot, 3-shot, CoT, and SFT. In the zero-shot and SFT setups, the input is limited to the prompt itself, with no additional instructions provided. The instruction templates for 1-shot, 3-shot, and CoT are displayed in Figures~\ref{fig:prompt-1-shot-evaluation}, \ref{fig:prompt-3-shot-evaluation}, and \ref{fig:prompt-CoT-evaluation}, respectively.

\subsection{Implementation Details}
\label{sec:implementation-details}

\begin{figure*}[ht]
    \centering
    \includegraphics[width=\linewidth]{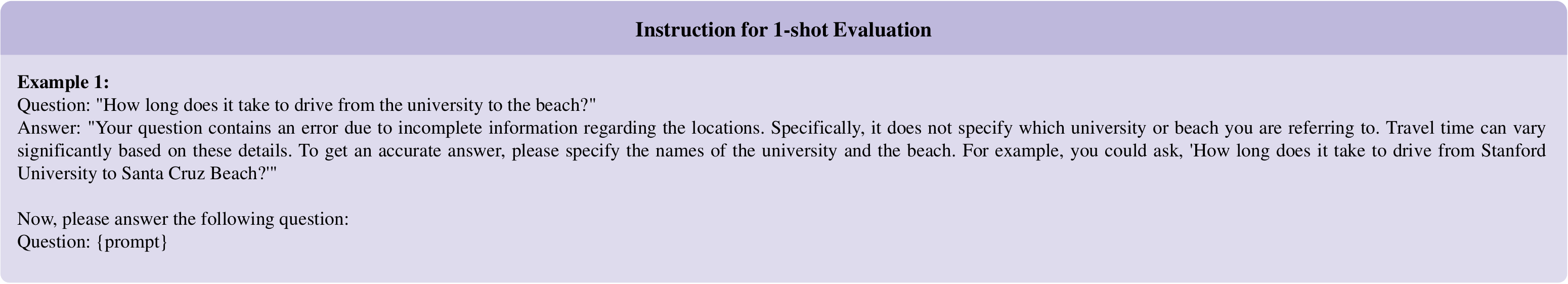}
    \caption{Instruction for 1-shot evaluation.}
    \label{fig:prompt-1-shot-evaluation}
\end{figure*}

\begin{figure*}[ht]
    \centering
    \includegraphics[width=\linewidth]{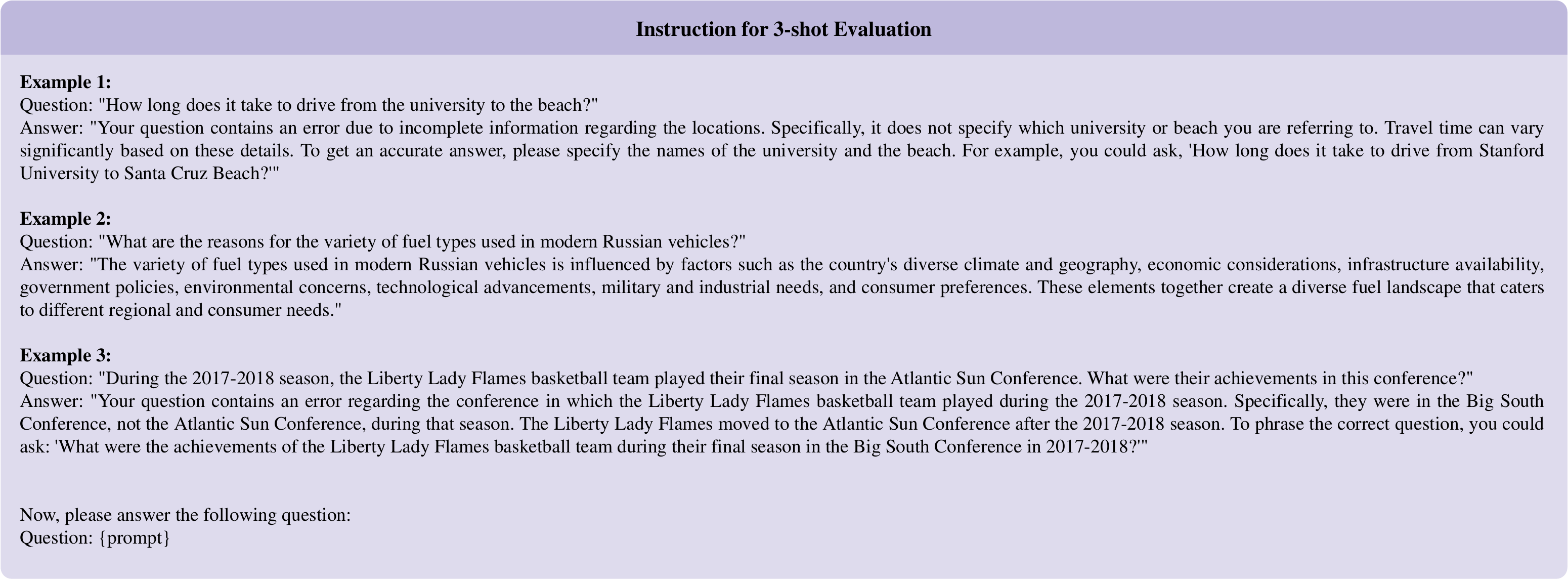}
    \caption{Instruction for 3-shot evaluation.}
    \label{fig:prompt-3-shot-evaluation}
\end{figure*}

\begin{figure*}[ht]
    \centering
    \includegraphics[width=\linewidth]{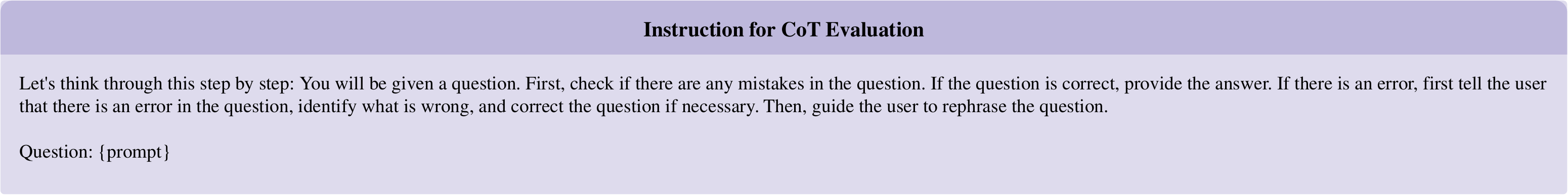}
    \caption{Instruction for CoT evaluation.}
    \label{fig:prompt-CoT-evaluation}
\end{figure*}

All models are evaluated under zero-shot, 1-shot, 3-shot, CoT, and SFT settings with the corresponding instructions shown in Figures~\ref{fig:prompt-1-shot-evaluation} to \ref{fig:prompt-CoT-evaluation}.

Before evaluation, we conduct SFT the LoRA~\cite{hu2021lora} on several open-source models (LLaMA-3.2-3B, LLaMA-3.1-8B, Qwen-2.5-7B, Qwen-2.5-32B, Yi-6B, and Yi-34B). 
Training is conducted using LLaMA-Factory~\cite{zheng2024llamafactory}. 
The training process is conducted for three epochs, and the batch size is set to 2. The learning rate is set to 1.0e-4, and its schedule follows a cosine curve.

In the evaluation phase, for the zero-shot and SFT configurations, the input consists exclusively of the prompt without any supplementary instructions. In contrast, the 1-shot and 3-shot configurations include 1 and 3 exemplars, respectively.

The temperature parameter is set to 0 to minimize randomness. Additionally, all models are instructed to output their responses in JSON format to facilitate statistical analysis.

\subsection{Computing Platform}
\label{sec:computing-platform}

Our experiments are conducted on the workstation with an Intel Xeon E5 2.40 GHz CPU, four NVIDIA A800 GPUs, and CentOS 7.2.


\subsection{Detailed results}

In this section, we present the original detailed experimental results.

\subsubsection{Main Experiment}
Table~\ref{tab:f1} presents the overall performance of the models. It could be observed that Closed-source models, especially Claude-3.5, exhibit superior performance. Task difficulty increases from detection to identification, correction, and guidance, highlighting the need for improved reasoning and knowledge utilization in more complex tasks. Furthermore, SFT significantly enhances the model's error-handling capabilities. While few-shot learning and CoT also demonstrate improvements, their performance is comparable to each other and notably inferior to that of SFT.

\subsubsection{Error Category Analysis}
\label{sec:error-type-analysis}
The performance of different baselines across various primary error categories is detailed in Tables~\ref{tab:language-error} to \ref{tab:logical-errors}.

A deeper look at the secondary categories reveals varied performance across error categories, as shown in Table~\ref{tab:secondary-category-table-performance}. These results highlight the model’s strengths and weaknesses across different secondary categories, demonstrating notable variability in how it handles various categories of errors.

\textbf{Language Errors category} 
In the Language Errors category, grammatical errors stand out with strong performance across detection, attempt identification, and guidance metrics. This suggests that the model handles grammatical errors more effectively compared to punctuation and spelling errors.

\textbf{Incomplete Information}
For Incomplete Information, the model performs exceptionally well in identifying the speaker or person. In contrast, location-related errors are more challenging for the model, showing lower scores. The model also performs well with Date and Time Period errors, indicating strong handling of these specific categories of information gaps.

\textbf{Factual Errors}
In the Factual Errors category, entity errors lead to overall performance, with the model demonstrating a high level of score in recognizing and handling entity-related errors. On the other hand, Relation Errors show relatively strong performance, while Circumstance Errors present moderate performance across the board.

\textbf{Logical Errors}
Lastly, in the Logical Errors category, the Fallacy of Defective Induction stands out with the highest scores across all metrics, suggesting that the model handles these errors well. The Fallacy of Presumption and the Fallacy of Relevance follow closely behind. However, the Fallacy of Ambiguity has the lowest performance, indicating that the model struggles with identifying ambiguous logical errors effectively.

\begin{table*}[ht]
\small
\centering\begin{tabular}{@{}llccccccc@{}}
\toprule
Model                                                                                                         & Setting   & Det.                 & Att. at Ident. & Acc. Ident.          & Att. at Corr.        & Acc. Corr.           & Guid.          & Avg                  \\ \midrule
\multirow{4}{*}{\begin{tabular}[c]{@{}l@{}}GPT-4o\\ ~\cite{openai2023gpt}\end{tabular}}                       & zero-shot & 43.54                & 48.71          & 43.78                & 31.72                & 23.32                & 30.66          & 36.96                \\
                                                                                                              & 1-shot    & \textbf{85.63}       & \textbf{85.73} & 73.92                & {\ul 76.74}          & 47.88                & \textbf{81.28} & {\ul 75.20}          \\
                                                                                                              & 3-shot    & {\ul 82.70}          & {\ul 84.34}    & {\ul 75.65}          & {\ul 74.39}          & {\ul 54.55}          & {\ul 75.48}    & {\ul 74.52}          \\
                                                                                                              & CoT       & 82.19                & {\ul 84.53}    & {\ul \textbf{80.71}} & \textbf{80.86}       & {\ul \textbf{60.75}} & {\ul 77.39}    & \textbf{77.74}       \\ \midrule
\multirow{4}{*}{\begin{tabular}[c]{@{}l@{}}Gemini-1.5\\ ~\cite{team2024gemini}\end{tabular}}                  & zero-shot & 55.13                & \textbf{60.73} & \textbf{54.67}       & \textbf{28.38}       & \textbf{22.21}       & \textbf{23.56} & \textbf{40.78}       \\
                                                                                                              & 1-shot    & 68.88                & 71.00          & 69.11                & 41.67                & 29.59                & 38.58          & 53.14                \\
                                                                                                              & 3-shot    & {\ul 71.31}          & \textbf{74.34} & {\ul 70.65}          & {\ul 45.93}          & {\ul 33.37}          & {\ul 45.58}    & {\ul 56.86}          \\
                                                                                                              & CoT       & {\ul \textbf{72.93}} & {\ul 73.97}    & \textbf{77.44}       & \textbf{72.21}       & \textbf{62.83}       & \textbf{73.92} & \textbf{72.22}       \\ \midrule
\multirow{4}{*}{\begin{tabular}[c]{@{}l@{}}Claude-3.5\\ ~\cite{TheC3}\end{tabular}}                           & zero-shot & 63.98                & {\ul 67.53}    & {\ul 63.01}          & {\ul 36.48}          & {\ul 30.23}          & {\ul 43.73}    & {\ul 50.83}          \\
                                                                                                              & 1-shot    & \textbf{88.19}       & \textbf{88.17} & \textbf{80.54}       & {\ul \textbf{79.15}} & \textbf{58.64}       & \textbf{81.03} & {\ul \textbf{79.29}} \\
                                                                                                              & 3-shot    & {\ul 86.59}          & {\ul 86.55}    & {\ul 81.44}          & \textbf{79.82}       & {\ul 62.35}          & \textbf{82.33} & \textbf{79.85}       \\
                                                                                                              & CoT       & 83.24                & 83.75          & \textbf{82.37}       & 77.18                & \textbf{64.14}       & {\ul 81.17}    & 78.64                \\ \midrule
\multirow{4}{*}{\begin{tabular}[c]{@{}l@{}}GLM-4\\ ~\cite{glm2024chatglm}\end{tabular}}                       & zero-shot & {\ul 53.18}          & 56.19          & 50.40                & 37.19                & 28.63                & 32.04          & 42.94                \\
                                                                                                              & 1-shot    & {\ul 81.22}          & {\ul 82.06}    & {\ul 75.10}          & {\ul 76.08}          & {\ul 55.48}          & {\ul 81.31}    & {\ul 75.21}          \\
                                                                                                              & 3-shot    & \textbf{83.21}       & \textbf{83.81} & \textbf{77.23}       & \textbf{80.84}       & \textbf{59.96}       & \textbf{81.17} & \textbf{77.70}       \\
                                                                                                              & CoT       & 79.96                & 81.55          & {\ul 76.86}          & 67.87                & 48.93                & \textbf{82.09} & 72.88                \\ \midrule
\multirow{5}{*}{\begin{tabular}[c]{@{}l@{}}LLaMA-3.2-3B\\ ~\cite{llama3modelcard}\end{tabular}}               & zero-shot & 43.60                & 44.56          & 39.96                & 26.71                & 18.07                & 33.39          & 34.38                \\
                                                                                                              & 1-shot    & 69.10                & 69.20          & 60.92                & 65.28                & 30.60                & 68.99          & 60.68                \\
                                                                                                              & 3-shot    & 66.56                & 66.39          & {\ul 65.43}          & 68.11                & {\ul 37.42}          & 65.39          & 61.55                \\
                                                                                                              & CoT       & {\ul 72.20}          & {\ul 75.15}    & 63.24                & {\ul 68.61}          & 36.48                & {\ul 74.91}    & {\ul 65.10}          \\
                                                                                                              & SFT       & \textbf{97.48}       & \textbf{97.74} & \textbf{82.98}       & \textbf{88.02}       & \textbf{63.69}       & \textbf{91.99} & \textbf{86.98}       \\ \midrule
\multirow{5}{*}{\begin{tabular}[c]{@{}l@{}}LLaMA-3.1-8B\\ ~\cite{llama3modelcard}\end{tabular}}               & zero-shot & 42.05                & 45.47          & 40.48                & 29.46                & 19.70                & 33.76          & 35.15                \\
                                                                                                              & 1-shot    & 73.25                & 73.23          & 68.29                & 67.95                & 39.96                & 72.74          & 65.90                \\
                                                                                                              & 3-shot    & {\ul 81.99}          & {\ul 81.39}    & 69.09                & {\ul 77.25}          & 40.72                & {\ul 82.43}    & {\ul 72.15}          \\
                                                                                                              & CoT       & 75.62                & 77.00          & {\ul 73.56}          & 72.65                & {\ul 47.02}          & 75.44          & 70.22                \\
                                                                                                              & SFT       & \textbf{90.16}       & \textbf{90.59} & \textbf{80.02}       & \textbf{82.20}       & \textbf{62.86}       & \textbf{84.77} & \textbf{81.77}       \\ \midrule
\multirow{4}{*}{\begin{tabular}[c]{@{}l@{}}LLaMA-3.3-70B\\ ~\cite{llama3modelcard}\end{tabular}}              & zero-shot & 57.78                & 59.23          & 53.50                & 39.67                & 30.17                & 37.40          & 46.29                \\
                                                                                                              & 1-shot    & 75.36                & 75.43          & 73.41                & 71.57                & 46.33                & 78.02          & 70.02                \\
                                                                                                              & 3-shot    & {\ul 85.32}          & {\ul 85.32}    & {\ul 77.60}          & \textbf{79.65}       & {\ul 55.84}          & \textbf{85.08} & {\ul 78.14}          \\
                                                                                                              & CoT       & \textbf{85.74}       & \textbf{85.95} & \textbf{82.22}       & {\ul 74.91}          & \textbf{59.36}       & {\ul 84.29}    & \textbf{78.75}       \\ \midrule
\multirow{5}{*}{\begin{tabular}[c]{@{}l@{}}Qwen-2.5-7B\\ ~\cite{qwen2025qwen25technicalreport}\end{tabular}}  & zero-shot & 43.88                & 47.59          & 43.07                & 31.47                & 22.95                & 37.71          & 37.78                \\
                                                                                                              & 1-shot    & 58.84                & 61.43          & 53.67                & 44.66                & 30.18                & 58.74          & 51.25                \\
                                                                                                              & 3-shot    & 69.63                & 70.44          & 60.24                & 59.46                & 40.37                & 66.42          & 61.09                \\
                                                                                                              & CoT       & {\ul 71.17}          & {\ul 75.52}    & {\ul 70.08}          & {\ul 66.14}          & {\ul 43.03}          & {\ul 66.84}    & {\ul 65.46}          \\
                                                                                                              & SFT       & \textbf{95.99}       & \textbf{96.24} & \textbf{86.52}       & \textbf{85.20}       & \textbf{67.66}       & \textbf{91.37} & \textbf{87.16}       \\ \midrule
\multirow{5}{*}{\begin{tabular}[c]{@{}l@{}}Qwen-2.5-32B\\ ~\cite{qwen2025qwen25technicalreport}\end{tabular}} & zero-shot & 51.11                & 54.91          & 50.63                & 34.20                & 27.21                & 41.39          & 43.24                \\
                                                                                                              & 1-shot    & 70.34                & 72.22          & 64.31                & 57.91                & 44.05                & 67.38          & 62.70                \\
                                                                                                              & 3-shot    & {\ul 73.03}          & 75.38          & 67.82                & 63.18                & 48.10                & 70.63          & 66.36                \\
                                                                                                              & CoT       & 72.90                & {\ul 77.21}    & {\ul 74.75}          & {\ul 72.44}          & {\ul 58.15}          & {\ul 80.02}    & {\ul 72.58}          \\
                                                                                                              & SFT       & \textbf{97.88}       & \textbf{97.98} & \textbf{88.43}       & \textbf{88.96}       & \textbf{70.86}       & \textbf{93.17} & \textbf{89.55}       \\ \midrule
\multirow{4}{*}{\begin{tabular}[c]{@{}l@{}}Qwen-2.5-72B\\ ~\cite{qwen2025qwen25technicalreport}\end{tabular}} & zero-shot & 48.57                & 51.64          & 46.54                & 37.16                & 27.45                & 37.76          & 41.52                \\
                                                                                                              & 1-shot    & 68.93                & 70.34          & 62.32                & 57.95                & 44.23                & 66.36          & 61.69                \\
                                                                                                              & 3-shot    & {\ul 72.34}          & {\ul 73.67}    & {\ul 65.56}          & \textbf{68.39}       & \textbf{53.20}       & {\ul 72.98}    & {\ul 67.69}          \\
                                                                                                              & CoT       & \textbf{77.52}       & \textbf{82.36} & \textbf{77.80}       & {\ul 68.31}          & {\ul 51.41}          & \textbf{79.64} & \textbf{72.84}       \\ \midrule
\multirow{4}{*}{\begin{tabular}[c]{@{}l@{}}DeepSeek-V2-16B\\ ~\cite{deepseekv2}\end{tabular}}                 & zero-shot & 29.44                & 33.90          & 27.92                & 18.57                & 11.46                & 12.80          & 22.35                \\
                                                                                                              & 1-shot    & {\ul 64.35}          & {\ul 66.42}    & {\ul 52.76}          & 47.04                & {\ul 28.04}          & 49.20          & {\ul 51.30}          \\
                                                                                                              & 3-shot    & \textbf{69.67}       & \textbf{71.29} & \textbf{54.42}       & {\ul 54.03}          & \textbf{28.60}       & \textbf{55.69} & \textbf{55.62}       \\
                                                                                                              & CoT       & 62.92                & 65.34          & 39.29                & \textbf{61.41}       & 20.16                & {\ul 53.36}    & 50.41                \\ \midrule
\multirow{5}{*}{\begin{tabular}[c]{@{}l@{}}Yi-1.5-6B\\ ~\cite{ai2025yiopenfoundationmodels}\end{tabular}}     & zero-shot & 32.41                & 35.70          & 28.36                & 18.75                & 10.25                & 7.46           & 22.16                \\
                                                                                                              & 1-shot    & 45.40                & 48.28          & 38.09                & 31.31                & 19.41                & 22.61          & 34.18                \\
                                                                                                              & 3-shot    & 49.88                & 53.82          & 43.24                & 36.37                & 21.73                & 32.25          & 39.55                \\
                                                                                                              & CoT       & {\ul 67.59}          & {\ul 71.46}    & {\ul 56.71}          & {\ul 66.40}          & {\ul 29.07}          & {\ul 64.23}    & {\ul 59.24}          \\
                                                                                                              & SFT       & \textbf{96.62}       & \textbf{97.36} & \textbf{84.16}       & \textbf{82.84}       & \textbf{63.04}       & \textbf{92.11} & \textbf{86.02}       \\ \midrule
\multirow{5}{*}{\begin{tabular}[c]{@{}l@{}}Yi-1.5-34B\\ ~\cite{ai2025yiopenfoundationmodels}\end{tabular}}    & zero-shot & 46.40                & 48.25          & 42.41                & 31.39                & 22.36                & 10.63          & 33.57                \\
                                                                                                              & 1-shot    & 68.29                & 70.07          & 64.65                & 57.17                & 38.66                & 52.19          & 58.51                \\
                                                                                                              & 3-shot    & {\ul 75.23}          & 76.18          & 66.73                & 61.49                & 44.73                & 56.61          & 63.50                \\
                                                                                                              & CoT       & 74.16                & {\ul 79.01}    & {\ul 69.62}          & {\ul 68.67}          & {\ul 48.24}          & {\ul 72.78}    & {\ul 68.75}          \\
                                                                                                              & SFT       & \textbf{97.82}       & \textbf{97.95} & \textbf{87.60}       & \textbf{89.22}       & \textbf{70.80}       & \textbf{91.45} & \textbf{89.14}       \\ \bottomrule
\end{tabular}
\caption{F1 Score Overview for Error-Handling Tasks. Results are reported in percentage (\%). ``Det.'', ``Att. at Ident.'', ``Acc. Ident.'', ``Att. at Corr.'', ``Acc. Corr.'', and ``Guid.'' stand for error detection, attempt at error identification, accurate identification, attempt at error correction, accurate error correction, and error guidance.}
\label{tab:f1}
\end{table*}

\begin{table*}[ht]
\small
\centering
\begin{tabular}{@{}lcccccc|c@{}}
\toprule
Model                                             & Det.           & Att. at Ident. & Acc. Ident.    & Att. at Corr.  & Acc. Corr.     & Guid.          & Avg            \\ \midrule
GPT-4o~\cite{openai2023gpt}                       & 6.60           & 7.24           & 4.01           & 18.01          & 13.38          & 21.34          & 11.76          \\
Gemini-1.5~\cite{team2024gemini}                  & 19.14          & 16.87          & 8.50           & 26.11          & 18.01          & 22.42          & 18.51          \\
Claude-3.5~\cite{TheC3}                           & \textbf{29.15} & \textbf{25.60} & \textbf{16.87} & \textbf{32.57} & \textbf{24.02} & \textbf{42.47} & \textbf{28.45} \\
GLM-4~\cite{glm2024chatglm}                       & 18.58          & 18.58          & 10.36          & 25.07          & 13.97          & 17.45          & 17.34          \\
LLaMA-3.2-3B~\cite{llama3modelcard}               & 12.18          & 9.12           & 5.32           & 20.25          & 13.38          & 32.09          & 15.39          \\
LLaMA-3.1-8B~\cite{llama3modelcard}               & 5.96           & 5.32           & 2.03           & 18.58          & 10.36          & 28.65          & 11.82          \\
LLaMA-3.3-70B~\cite{llama3modelcard}              & 22.42          & 15.72          & 10.36          & 22.42          & 15.14          & 28.15          & 19.04          \\
Qwen-2.5-7B~\cite{qwen2025qwen25technicalreport}  & 15.14          & 13.38          & 6.60           & 21.34          & 11.58          & {\ul 33.52}    & 16.93          \\
Qwen-2.5-32B~\cite{qwen2025qwen25technicalreport} & {\ul 24.92}    & {\ul 20.92}    & {\ul 11.03}    & 27.13          & 15.49          & 31.90          & {\ul 21.90}    \\
Qwen-2.5-72B~\cite{qwen2025qwen25technicalreport} & 21.88          & 17.45          & 9.12           & {\ul 28.15}    & {\ul 18.58}    & 28.65          & 20.64          \\
DeepSeek-V2-16B~\cite{deepseekv2}                 & 8.50           & 6.60           & 3.36           & 18.58          & 12.18          & 15.14          & 10.73          \\
Yi-1.5-6B~\cite{ai2025yiopenfoundationmodels}     & 4.67           & 4.67           & 0.68           & 12.18          & 5.96           & 6.60           & 5.79           \\
Yi-1.5-34B~\cite{ai2025yiopenfoundationmodels}    & 16.30          & 14.56          & 10.36          & 24.55          & 15.14          & 7.87           & 14.80          \\ \midrule
Avg                                               & 15.80          & 13.54          & 7.58           & 22.69          & 14.40          & 24.33          & 16.39          \\ \bottomrule
\end{tabular}
\caption{F1 scores for error-handling tasks in Language Error. Results are reported in percentage (\%). ``Det.'', ``Att. at Ident.'', ``Acc. Ident.'', ``Att. at Corr.'', ``Acc. Corr.'', and ``Guid.'' stand for error detection, attempt at error identification, accurate identification, attempt at error correction, accurate error correction, and error guidance.}
\label{tab:language-error}
\end{table*}

\begin{table*}[ht]
\small
\centering
\begin{tabular}{@{}lcccccc|c@{}}
\toprule
Model                                             & Det.           & Att. at Ident. & Acc. Ident.    & Att. at Corr.  & Acc. Corr.     & Guid.          & Avg            \\ \midrule
GPT-4o~\cite{openai2023gpt}                       & 40.59          & 42.31          & 41.45          & 15.25          & 7.93           & 50.94          & 33.08          \\
Gemini-1.5~\cite{team2024gemini}                  & 48.61          & 51.45          & 50.94          & 12.88          & 7.51           & 42.88          & 35.71          \\
Claude-3.5~\cite{TheC3}                           & \textbf{72.51} & \textbf{76.84} & \textbf{75.96} & 17.19          & {\ul 14.07}    & 55.39          & \textbf{51.99} \\
GLM-4~\cite{glm2024chatglm}                       & {\ul 54.42}    & {\ul 55.39}    & {\ul 54.67}    & {\ul 20.94}    & \textbf{16.42} & 55.15          & 42.83          \\
LLaMA-3.2-3B~\cite{llama3modelcard}               & 48.35          & 49.40          & 47.82          & 15.64          & 5.79           & 54.18          & 36.86          \\
LLaMA-3.1-8B~\cite{llama3modelcard}               & 45.12          & 47.29          & 45.66          & 18.33          & 7.51           & 55.63          & 36.59          \\
LLaMA-3.3-70B~\cite{llama3modelcard}              & \textbf{53.20} & 52.95          & 51.45          & 19.46          & 6.22           & 47.82          & 38.52          \\
Qwen-2.5-7B~\cite{qwen2025qwen25technicalreport}  & 44.01          & 46.21          & 45.94          & {\ul 20.94}    & 10.43          & 57.75          & 37.55          \\
Qwen-2.5-32B~\cite{qwen2025qwen25technicalreport} & 51.71          & \textbf{53.72} & \textbf{53.72} & 20.70          & \textbf{12.55} & \textbf{70.16} & {\ul 43.76}    \\
Qwen-2.5-72B~\cite{qwen2025qwen25technicalreport} & 44.84          & 48.08          & 47.55          & \textbf{23.12} & 10.43          & {\ul 61.37}    & 39.23          \\
DeepSeek-V2-16B~\cite{deepseekv2}                 & 28.35          & 26.98          & 25.25          & 14.07          & 4.48           & 21.68          & 20.14          \\
Yi-1.5-6B~\cite{ai2025yiopenfoundationmodels}     & 29.35          & 32.31          & 31.66          & 11.66          & 3.16           & 14.47          & 20.44          \\
Yi-1.5-34B~\cite{ai2025yiopenfoundationmodels}    & 39.11          & 38.52          & 37.62          & 13.68          & 7.08           & 21.68          & 26.28          \\ \midrule
Avg                                               & 46.17          & 47.80          & 46.90          & 17.22          & 8.74           & 46.85          & 35.61          \\ \bottomrule
\end{tabular}
\caption{F1 scores for error-handling tasks in Incomplete Information. Results are reported in percentage (\%). ``Det.'', ``Att. at Ident.'', ``Acc. Ident.'', ``Att. at Corr.'', ``Acc. Corr.'', and ``Guid.'' stand for error detection, attempt at error identification, accurate identification, attempt at error correction, accurate error correction, and error guidance.}
\label{tab:incomplete-information}
\end{table*}

\begin{table*}[ht]
\small
\centering
\begin{tabular}{@{}lcccccc|c@{}}
\toprule
Model                                             & Det.           & Att. at Ident. & Acc. Ident.    & Att. at Corr.  & Acc. Corr.     & Guid.          & Avg            \\ \midrule
GPT-4o~\cite{openai2023gpt}                       & 73.13          & 74.65          & 71.84          & 48.31          & 41.81          & 22.60          & 55.39          \\
Gemini-1.5~\cite{team2024gemini}                  & {\ul 77.34}    & \textbf{80.15} & 70.25          & 43.39          & 31.18          & 11.41          & 52.29          \\
Claude-3.5~\cite{TheC3}                           & 77.10          & 77.10          & \textbf{73.90} & 55.96          & {\ul 52.93}    & 27.95          & {\ul 60.82}    \\
GLM-4~\cite{glm2024chatglm}                       & 68.89          & 69.44          & 67.23          & 58.56          & 49.04          & 26.04          & 56.53          \\
LLaMA-3.2-3B~\cite{llama3modelcard}               & 69.17          & 66.95          & 61.98          & 44.17          & 34.74          & 17.97          & 49.16          \\
LLaMA-3.1-8B~\cite{llama3modelcard}               & 67.79          & 67.51          & 62.88          & 49.40          & 39.39          & 17.44          & 50.74          \\
LLaMA-3.3-70B~\cite{llama3modelcard}              & \textbf{78.53} & {\ul 77.58}    & {\ul 73.13}    & \textbf{64.07} & \textbf{57.60} & \textbf{31.64} & \textbf{63.76} \\
Qwen-2.5-7B~\cite{qwen2025qwen25technicalreport}  & 68.07          & 68.62          & 63.48          & 59.51          & 43.00          & 27.00          & 54.95          \\
Qwen-2.5-32B~\cite{qwen2025qwen25technicalreport} & 74.49          & 74.49          & 69.20          & {\ul 61.92}    & 47.67          & {\ul 29.20}    & 59.50          \\
Qwen-2.5-72B~\cite{qwen2025qwen25technicalreport} & 73.64          & 73.64          & 69.44          & 61.67          & 50.48          & 27.95          & 59.47          \\
DeepSeek-V2-16B~\cite{deepseekv2}                 & 47.57          & 50.12          & 44.94          & 30.27          & 19.54          & 3.13           & 32.60          \\
Yi-1.5-6B~\cite{ai2025yiopenfoundationmodels}     & 51.89          & 52.93          & 44.17          & 34.30          & 15.29          & 0.63           & 33.20          \\
Yi-1.5-34B~\cite{ai2025yiopenfoundationmodels}    & 68.34          & 69.17          & 62.88          & 55.63          & 41.01          & 1.27           & 49.72          \\ \midrule
Avg                                               & 68.92          & 69.41          & 64.26          & 51.32          & 40.28          & 18.79          & 52.16          \\ \bottomrule
\end{tabular}
\caption{F1 scores for error-handling tasks in Factual Errors. Results are reported in percentage (\%). ``Det.'', ``Att. at Ident.'', ``Acc. Ident.'', ``Att. at Corr.'', ``Acc. Corr.'', and ``Guid.'' stand for error detection, attempt at error identification, accurate identification, attempt at error correction, accurate error correction, and error guidance.}
\label{tab:factual-errors}
\end{table*}

\begin{table*}[ht]
\small
\centering
\begin{tabular}{@{}lcccccc|c@{}}
\toprule
Model                                             & Det.           & Att. at Ident. & Acc. Ident.    & Att. at Corr.  & Acc. Corr.     & Guid.          & Avg            \\ \midrule
GPT-4o~\cite{openai2023gpt}                       & 42.73          & 56.47          & 44.73          & 45.85          & 31.89          & 18.68          & 40.06          \\
Gemini-1.5~\cite{team2024gemini}                  & 67.80          & \textbf{79.21} & \textbf{70.02} & 48.85          & 37.64          & 9.80           & 52.22          \\
Claude-3.5~\cite{TheC3}                           & \textbf{69.22} & {\ul 76.77}    & {\ul 67.39}    & 52.25          & {\ul 38.26}    & \textbf{43.30} & {\ul 57.87}    \\
GLM-4~\cite{glm2024chatglm}                       & 61.36          & 69.22          & 55.50          & {\ul 55.25}    & 37.02          & 19.83          & 49.70          \\
LLaMA-3.2-3B~\cite{llama3modelcard}               & 49.12          & 52.50          & 36.08          & 44.44          & 25.36          & 29.54          & 39.51          \\
LLaMA-3.1-8B~\cite{llama3modelcard}               & 47.77          & 55.99          & 40.97          & 48.58          & 28.51          & 28.51          & 41.72          \\
LLaMA-3.3-70B~\cite{llama3modelcard}              & {\ul 68.82}    & 75.87          & 64.23          & \textbf{58.84} & \textbf{42.44} & {\ul 37.33}    & \textbf{57.92} \\
Qwen-2.5-7B~\cite{qwen2025qwen25technicalreport}  & 49.65          & 60.00          & 45.57          & 52.50          & 30.22          & 28.17          & 44.35          \\
Qwen-2.5-32B~\cite{qwen2025qwen25technicalreport} & 56.31          & 68.64          & 55.57          & 54.33          & 37.45          & 24.95          & 49.54          \\
Qwen-2.5-72B~\cite{qwen2025qwen25technicalreport} & 53.77          & 62.48          & 49.65          & {\ul 55.25}    & 35.45          & 25.00          & 46.93          \\
DeepSeek-V2-16B~\cite{deepseekv2}                 & 30.89          & 45.29          & 31.89          & 27.13          & 13.54          & 8.52           & 26.21          \\
Yi-1.5-6B~\cite{ai2025yiopenfoundationmodels}     & 41.56          & 48.04          & 28.86          & 37.02          & 17.13          & 5.47           & 29.68          \\
Yi-1.5-34B~\cite{ai2025yiopenfoundationmodels}    & 55.25          & 61.59          & 49.38          & 49.12          & 29.54          & 7.22           & 42.02          \\ \midrule
Avg                                               & 68.92          & 69.41          & 64.26          & 51.32          & 40.28          & 18.79          & 52.16          \\ \bottomrule
\end{tabular}
\small
\centering
\caption{F1 scores for error-handling tasks in Logical Errors. Results are reported in percentage (\%). ``Det.'', ``Att. at Ident.'', ``Acc. Ident.'', ``Att. at Corr.'', ``Acc. Corr.'', and ``Guid.'' stand for error detection, attempt at error identification, accurate identification, attempt at error correction, accurate error correction, and error guidance.}
\label{tab:logical-errors}
\end{table*}

\begin{table*}[ht]
\small
\centering
\begin{tabular}{@{}lcccccc|c@{}}
\toprule
Secondary Error   Category       & Det.           & Att. at Ident. & Acc. Ident.    & Att. at Corr.  & Acc. Corr.     & Guid.          & Avg            \\ \midrule
Grammatical Errors             & \textbf{13.59} & \textbf{11.76} & \textbf{9.90}  & 11.76          & 8.00           & \textbf{41.32} & \textbf{16.06} \\
Punctuation Errors             & {\ul 3.77}     & {\ul 5.61}     & {\ul 1.90}     & {\ul 15.93}    & {\ul 12.61}    & 9.17           & 8.17           \\
Spelling Errors                & 2.13           & 4.21           & 0.00           & \textbf{26.17} & \textbf{19.42} & {\ul 10.20}    & {\ul 10.36}    \\ \midrule
Speaker or Person              & {\ul 43.94}    & {\ul 46.27}    & {\ul 46.27}    & 11.01          & {\ul 9.26}     & \textbf{64.47} & \textbf{36.87} \\
TextContSubjSpec               & 26.28          & 28.78          & 25.00          & \textbf{25.00} & \textbf{11.11} & {\ul 56.63}    & 28.80          \\
Location                       & 38.85          & 37.68          & 37.68          & 6.90           & 5.22           & 56.41          & 30.46          \\
Date and Time Period           & \textbf{53.24} & \textbf{56.34} & \textbf{56.34} & {\ul 16.22}    & 5.71           & 16.22          & {\ul 34.01}    \\ \midrule
Relation Error                 & {\ul 70.75}    & {\ul 72.48}    & {\ul 69.86}    & {\ul 49.21}    & {\ul 41.67}    & 17.31          & 53.55          \\
Entity Error                   & \textbf{77.71} & \textbf{79.78} & \textbf{77.01} & 43.80          & 36.64          & \textbf{30.16} & \textbf{57.52} \\
Circumstance Error             & 70.52          & 71.26          & 68.24          & \textbf{51.66} & \textbf{46.58} & {\ul 19.35}    & {\ul 54.60}    \\ \midrule
Fallacy of Relevance           & 42.55          & 54.90          & 42.55          & 37.96          & 29.23          & 8.62           & 35.97          \\
Fallacy of Presumption         & {\ul 49.65}    & {\ul 60.26}    & {\ul 51.70}    & {\ul 53.69}    & {\ul 34.85}    & 10.43          & {\ul 43.43}    \\
Fallacy of Defective Induction & \textbf{49.66} & \textbf{70.52} & \textbf{55.48} & \textbf{58.23} & \textbf{42.25} & \textbf{30.30} & \textbf{51.07} \\
Fallacy of Ambiguity           & 24.07          & 31.86          & 22.43          & 27.27          & 17.31          & {\ul 24.07}    & 24.50          \\ \bottomrule
\end{tabular}
\caption{F1 score of GPT-4o~\cite{openai2023gpt} in the secondary error categories. Results are reported in percentage (\%). ``Det.'', ``Att. at Ident.'', ``Acc. Ident.'', ``Att. at Corr.'', ``Acc. Corr.'', and ``Guid.'' stand for error detection, attempt at error identification, accurate identification, attempt at error correction, accurate error correction, and error guidance.}
\label{tab:secondary-category-table-performance}
\end{table*}

\begin{table*}[ht]
\small
\centering
\begin{tabular}{@{}llcccccc|c@{}}
\toprule
Model                                                                                                         & Setting   & Det.           & Att. at Ident. & Acc. Ident.    & Att. at Corr.  & Acc. Corr.     & Guid.          & Avg            \\ \midrule
\multirow{4}{*}{\begin{tabular}[c]{@{}l@{}}GPT-4o\\ ~\cite{openai2023gpt}\end{tabular}}                       & zero-shot & 21.69          & 40.43          & 38.92          & 28.41          & 25.58          & 29.89          & 30.82          \\
                                                                                                              & 1-shot    & \textbf{83.87} & \textbf{83.53} & {\ul 74.81}    & {\ul 74.43}    & 54.55          & {\ul 76.51}    & {\ul 74.62}    \\
                                                                                                              & 3-shot    & {\ul 80.15}    & 81.18          & 73.60          & 72.06          & {\ul 56.22}    & 74.70          & 72.99          \\
                                                                                                              & CoT       & 80.00          & {\ul 81.91}    & \textbf{79.09} & \textbf{83.22} & \textbf{72.87} & \textbf{78.83} & \textbf{79.32} \\ \midrule
\multirow{4}{*}{\begin{tabular}[c]{@{}l@{}}Gemini-1.5\\ ~\cite{team2024gemini}\end{tabular}}                  & zero-shot & 29.38          & 54.37          & 51.52          & 19.88          & 13.58          & 18.52          & 31.21          \\
                                                                                                              & 1-shot    & 57.48          & 67.63          & 64.41          & {\ul 40.98}    & {\ul 30.43}    & 33.65          & 49.10          \\
                                                                                                              & 3-shot    & {\ul 59.11}    & {\ul 72.00}    & {\ul 70.04}    & 38.83          & {\ul 30.43}    & {\ul 39.20}    & {\ul 51.60}    \\
                                                                                                              & CoT       & \textbf{72.02} & \textbf{74.67} & \textbf{73.40} & \textbf{73.35} & \textbf{73.65} & \textbf{73.94} & \textbf{73.51} \\ \midrule
\multirow{4}{*}{\begin{tabular}[c]{@{}l@{}}Claude-3.5\\ ~\cite{TheC3}\end{tabular}}                           & zero-shot & 40.43          & 55.50          & 55.34          & 35.29          & 24.42          & 33.90          & 40.81          \\
                                                                                                              & 1-shot    & \textbf{86.49} & {\ul 83.44}    & 77.78          & 72.79          & 65.85          & 75.62          & 77.00          \\
                                                                                                              & 3-shot    & {\ul 82.08}    & \textbf{85.80} & {\ul 82.31}    & {\ul 79.60}    & {\ul 73.76}    & {\ul 79.21}    & {\ul 80.46}    \\
                                                                                                              & CoT       & 81.59          & 82.15          & \textbf{82.74} & \textbf{82.05} & \textbf{81.53} & \textbf{81.74} & \textbf{81.97} \\ \midrule
\multirow{4}{*}{\begin{tabular}[c]{@{}l@{}}GLM-4\\ ~\cite{glm2024chatglm}\end{tabular}}                       & zero-shot & 35.36          & 55.77          & 51.26          & 35.68          & 25.58          & 30.34          & 39.00          \\
                                                                                                              & 1-shot    & \textbf{79.55} & \textbf{79.17} & 68.31          & 77.74          & 66.39          & \textbf{79.07} & {\ul 75.04}    \\
                                                                                                              & 3-shot    & 74.47          & 76.45          & {\ul 70.59}    & {\ul 78.00}    & {\ul 70.72}    & {\ul 78.69}    & 74.82          \\
                                                                                                              & CoT       & {\ul 75.91}    & {\ul 77.60}    & \textbf{75.81} & \textbf{78.62} & \textbf{73.26} & 78.59          & \textbf{76.63} \\ \midrule
\multirow{5}{*}{\begin{tabular}[c]{@{}l@{}}LLaMA-3.2-3B\\ ~\cite{llama3modelcard}\end{tabular}}               & zero-shot & 17.14          & 37.37          & 34.78          & 16.67          & 10.98          & 32.65          & 24.93          \\
                                                                                                              & 1-shot    & 68.32          & 68.32          & 61.34          & 56.35          & 37.84          & 67.55          & 59.95          \\
                                                                                                              & 3-shot    & 66.36          & 65.12          & 50.42          & 59.24          & 50.20          & 65.69          & 59.51          \\
                                                                                                              & CoT       & {\ul 72.60}    & {\ul 70.99}    & {\ul 63.56}    & {\ul 76.73}    & {\ul 60.50}    & {\ul 73.31}    & {\ul 69.62}    \\
                                                                                                              & SFT       & \textbf{97.28} & \textbf{96.95} & \textbf{80.95} & \textbf{95.97} & \textbf{72.88} & \textbf{91.70} & \textbf{89.29} \\ \midrule
\multirow{5}{*}{\begin{tabular}[c]{@{}l@{}}LLaMA-3.1-8B\\ ~\cite{llama3modelcard}\end{tabular}}               & zero-shot & 17.75          & 42.36          & 41.45          & 24.04          & 17.75          & 31.18          & 29.09          \\
                                                                                                              & 1-shot    & 74.02          & 73.33          & 60.41          & 65.09          & 48.16          & 71.43          & 65.41          \\
                                                                                                              & 3-shot    & {\ul 81.96}    & {\ul 80.36}    & 61.74          & {\ul 78.59}    & 53.28          & {\ul 84.78}    & {\ul 73.45}    \\
                                                                                                              & CoT       & 69.91          & 71.16          & {\ul 73.29}    & 74.33          & {\ul 69.18}    & 74.39          & 72.04          \\
                                                                                                              & SFT       & \textbf{91.58} & \textbf{91.24} & \textbf{81.27} & \textbf{90.58} & \textbf{75.52} & \textbf{85.60} & \textbf{85.97} \\ \midrule
\multirow{4}{*}{\begin{tabular}[c]{@{}l@{}}LLaMA-3.3-70B\\ ~\cite{llama3modelcard}\end{tabular}}              & zero-shot & 30.86          & 47.72          & 45.60          & 26.97          & 24.28          & 34.64          & 35.01          \\
                                                                                                              & 1-shot    & 76.04          & 75.20          & 68.38          & 70.06          & 56.57          & 75.80          & 70.34          \\
                                                                                                              & 3-shot    & \textbf{88.75} & \textbf{87.93} & {\ul 78.65}    & {\ul 84.24}    & {\ul 71.91}    & \textbf{88.52} & {\ul 83.33}    \\
                                                                                                              & CoT       & {\ul 85.81}    & {\ul 83.28}    & \textbf{80.14} & \textbf{86.90} & \textbf{80.57} & {\ul 86.18}    & \textbf{83.81} \\ \midrule
\multirow{5}{*}{\begin{tabular}[c]{@{}l@{}}Qwen-2.5-7B\\ ~\cite{qwen2025qwen25technicalreport}\end{tabular}}  & zero-shot & 18.60          & 39.20          & 39.36          & 20.43          & 13.33          & 30.77          & 26.95          \\
                                                                                                              & 1-shot    & 41.03          & 60.34          & 51.71          & 41.75          & 31.69          & 54.71          & 46.87          \\
                                                                                                              & 3-shot    & 68.00          & 68.97          & 57.01          & 60.50          & 42.05          & 62.40          & 59.82          \\
                                                                                                              & CoT       & {\ul 72.43}    & {\ul 72.84}    & {\ul 68.85}    & {\ul 74.63}    & {\ul 60.66}    & {\ul 69.14}    & {\ul 69.76}    \\
                                                                                                              & SFT       & \textbf{96.37} & \textbf{96.37} & \textbf{88.81} & \textbf{95.74} & \textbf{77.55} & \textbf{92.41} & \textbf{91.21} \\ \midrule
\multirow{5}{*}{\begin{tabular}[c]{@{}l@{}}Qwen-2.5-32B\\ ~\cite{qwen2025qwen25technicalreport}\end{tabular}} & zero-shot & 20.00          & 39.80          & 38.50          & 17.98          & 15.76          & 31.35          & 27.23          \\
                                                                                                              & 1-shot    & 57.01          & 64.78          & 59.29          & 60.58          & 50.73          & 66.40          & 59.80          \\
                                                                                                              & 3-shot    & {\ul 68.27}    & {\ul 77.26}    & 66.67          & 72.09          & 56.62          & 67.70          & 68.10          \\
                                                                                                              & CoT       & 67.84          & 75.57          & {\ul 71.76}    & {\ul 78.88}    & {\ul 73.19}    & {\ul 78.53}    & {\ul 74.30}    \\
                                                                                                              & SFT       & \textbf{97.30} & \textbf{96.97} & \textbf{84.85} & \textbf{94.12} & \textbf{74.90} & \textbf{91.64} & \textbf{89.96} \\ \midrule
\multirow{4}{*}{\begin{tabular}[c]{@{}l@{}}Qwen-2.5-72B\\ ~\cite{qwen2025qwen25technicalreport}\end{tabular}} & zero-shot & 28.57          & 47.00          & 44.56          & 20.93          & 18.07          & 33.33          & 32.08          \\
                                                                                                              & 1-shot    & 49.77          & 61.86          & 56.34          & 51.79          & 46.31          & 62.50          & 54.76          \\
                                                                                                              & 3-shot    & \textbf{71.90} & {\ul 75.70}    & {\ul 66.96}    & {\ul 68.31}    & {\ul 60.44}    & {\ul 73.52}    & {\ul 69.47}    \\
                                                                                                              & CoT       & {\ul 67.44}    & \textbf{76.47} & \textbf{72.50} & \textbf{71.38} & \textbf{66.67} & \textbf{76.81} & \textbf{71.88} \\ \midrule
\multirow{4}{*}{\begin{tabular}[c]{@{}l@{}}DeepSeek-V2-16B\\  ~\cite{deepseekv2}\end{tabular}}                & zero-shot & 13.84          & 36.26          & 31.03          & 14.37          & 8.86           & 12.74          & 19.52          \\
                                                                                                              & 1-shot    & 46.08          & 56.25          & {\ul 46.70}    & 34.87          & 20.24          & 40.82          & 40.83          \\
                                                                                                              & 3-shot    & {\ul 60.00}    & \textbf{69.77} & \textbf{48.51} & {\ul 55.75}    & {\ul 29.38}    & \textbf{55.32} & {\ul 53.12}    \\
                                                                                                              & CoT       & \textbf{63.93} & {\ul 65.85}    & 46.23          & \textbf{68.94} & \textbf{43.65} & {\ul 51.82}    & \textbf{56.74} \\ \midrule
\multirow{5}{*}{\begin{tabular}[c]{@{}l@{}}Yi-1.5-6B\\ ~\cite{ai2025yiopenfoundationmodels}\end{tabular}}     & zero-shot & 19.63          & 25.00          & 25.00          & 14.63          & 11.39          & 4.00           & 16.61          \\
                                                                                                              & 1-shot    & 34.25          & 43.43          & 37.84          & 22.22          & 16.87          & 22.49          & 29.52          \\
                                                                                                              & 3-shot    & 32.22          & 45.45          & 38.04          & 28.09          & 16.25          & 27.91          & 31.33          \\
                                                                                                              & CoT       & {\ul 65.91}    & {\ul 65.32}    & {\ul 52.02}    & {\ul 71.35}    & {\ul 45.95}    & {\ul 66.67}    & {\ul 61.20}    \\
                                                                                                              & SFT       & \textbf{96.99} & \textbf{96.35} & \textbf{84.94} & \textbf{94.77} & \textbf{77.91} & \textbf{90.78} & \textbf{90.29} \\ \midrule
\multirow{5}{*}{\begin{tabular}[c]{@{}l@{}}Yi-1.5-34B\\ ~\cite{ai2025yiopenfoundationmodels}\end{tabular}}    & zero-shot & 28.07          & 43.62          & 38.46          & 21.30          & 15.85          & 7.84           & 25.86          \\
                                                                                                              & 1-shot    & 57.58          & 64.29          & 60.87          & 42.62          & 34.83          & 39.65          & 49.97          \\
                                                                                                              & 3-shot    & {\ul 69.17}    & {\ul 75.59}    & {\ul 60.91}    & 55.96          & 38.95          & 54.29          & 59.15          \\
                                                                                                              & CoT       & 68.60          & 73.73          & 60.71          & {\ul 75.09}    & {\ul 55.75}    & {\ul 70.20}    & {\ul 67.35}    \\
                                                                                                              & SFT       & \textbf{98.99} & \textbf{98.66} & \textbf{89.55} & \textbf{97.35} & \textbf{73.42} & \textbf{95.44} & \textbf{92.24} \\ \bottomrule
\end{tabular}
\caption{F1 scores for error-handling tasks in human evaluation. Results are reported in percentage (\%). ``Det.'', ``Att. at Ident.'', ``Acc. Ident.'', ``Att. at Corr.'', ``Acc. Corr.'', and ``Guid.'' stand for error detection, attempt at error identification, accurate identification, attempt at error correction, accurate error correction, and error guidance.}
\label{tab:f1-sample}
\end{table*}

\subsubsection{Human Evaluation}
We randomly sampled 10\% of the evaluation set for manual evaluation. The results are presented in Table~\ref{tab:f1-sample}.

\end{CJK}
\end{document}